\documentclass[10pt,journal,compsoc]{IEEEtran}

\usepackage{epsfig}
\usepackage{graphicx}
\usepackage{amsmath}
\usepackage{amssymb}
\usepackage{mathtools}
\usepackage{bbm}
\usepackage{makecell, verbatim, sidecap}
\usepackage{dsfont}
\usepackage{booktabs}

\usepackage[nocompress]{cite}
\usepackage[noend]{algpseudocode}
\usepackage{algorithmicx,algorithm}
\usepackage{balance}
\usepackage{multirow}
\usepackage[breaklinks=true,bookmarks=false]{hyperref}

\usepackage{color}

\renewcommand{\texttt}[1]{${{\tt #1}}$}

\usepackage{xspace}

\renewcommand{\vec}[1]{\ensuremath{\pmb{#1}}}
\newcommand{\mat}[1]{\ensuremath{\mathbf{#1}}}
\newcommand{\set}[1]{\ensuremath{\mathscr{#1}}}

\makeatletter
\@tfor\next:=abcdefghijklmnopqrstuvwxyxz\do
{\begingroup\edef\x{\endgroup
		\noexpand\@namedef{v\next}{\noexpand\vec{\next}}%
	}\x}
\@tfor\next:=ABCDEFGHIJKLMNOPQRSTUVWXYZ\do
{\begingroup\edef\x{\endgroup
		\noexpand\@namedef{m\next}{\noexpand\mat{\next}}%
	}\x}
\@tfor\next:=ABCDEFGHIJKLMNOPQRSTUVWXYZ\do
{\begingroup\edef\x{\endgroup
		\noexpand\@namedef{s\next}{\noexpand\set{\next}}%
	}\x}
\makeatother

\def\eg{{\it e.g.}\xspace}

\def\revise{\textcolor{black}}
\renewcommand{\paragraph}{\textbf}

\def\SexyName{{DyCo3D}\xspace}

\begin{document}
	
	\title{Dynamic Convolution for 3D Point Cloud Instance Segmentation}

	\author{
		Tong He, ~~
		Chunhua Shen, ~~
		Anton van den Hengel
		\thanks{
			T. He and A. van den Hengel are with
			The University of Adelaide, Australia.
			C. Shen is with Zhejiang University.
			C. Shen is the
			corresponding author  (e-mail: chunhua@me.com).
			\color{blue}
			Accepted to IEEE Trans. Pattern Analysis and Machine Intelligence (TPAMI).
			
		}
	}

	\IEEEtitleabstractindextext{
		
		\begin{abstract}
	In this paper, we come up with a simple yet effective approach for instance segmentation on 3D point cloud with strong robustness. Previous top-performing methods for this task adopt a bottom-up strategy, which often involves various inefficient operations or complex pipelines, such as grouping over-segmented components, introducing heuristic post-processing steps, and designing complex loss functions. 
	As a result, the inevitable variations of the instances sizes make it vulnerable and sensitive to the values of pre-defined hyper-parameters.
	To this end, we instead propose a novel pipeline that applies dynamic convolution to generate instance-aware parameters in response to the characteristics of the instances. 
	The representation capability of the parameters is greatly improved by gathering homogeneous points that have identical semantic categories and close votes for the geometric centroids. 
	Instances are then decoded via several simple convolution layers, where the parameters are generated depending on the input. 
	In addition, to introduce a large context and maintain limited computational overheads, a light-weight transformer is built upon the bottleneck layer to capture the long-range dependencies. With the only post-processing step, non-maximum suppression (NMS), we demonstrate a simpler and more robust approach that achieves promising performance on various datasets: ScanNetV2, S3DIS, and PartNet. The consistent improvements on both voxel- and point-based architectures imply the effectiveness of the proposed method. Code is available at: \url{https://git.io/DyCo3D}. 

\end{abstract}

	}

	\maketitle

	%
	\IEEEraisesectionheading{\section{Introduction}\label{sec:introduction}}

\begin{figure*}[htbp]
	\centering
	{
		\includegraphics[height=6.5cm]{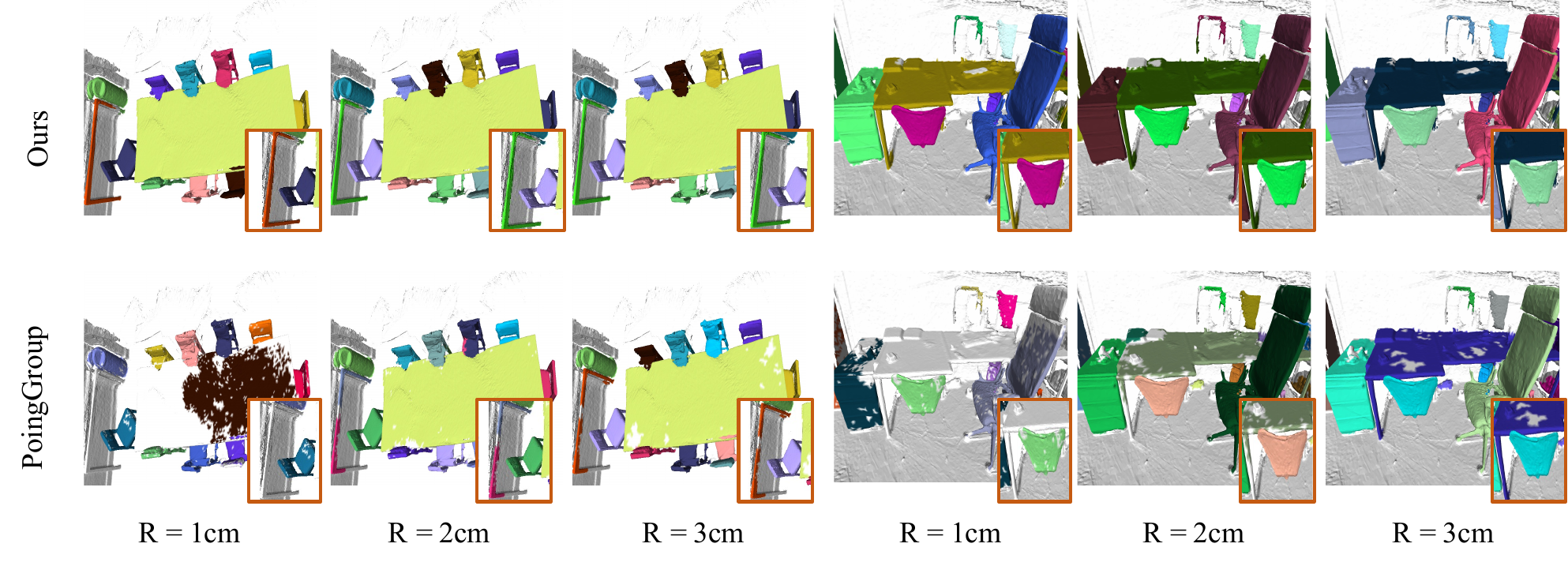}
	}
	\caption{A comparison of the instance segmentation results between our proposed \SexyName and one of the top-performing approach PointGroup \cite{jiang2020pointgroup}. Our method shows strong robustness and generalization capability to the varying values of cluster radius, which is one of the most important hyper-parameters to the success of \cite{jiang2020pointgroup}. Different instances are present with random colors, and rectangles highlight specific over-segmentation errors.  Best viewed in color.
	}
	\label{fig:main}
\end{figure*}

\IEEEPARstart{I}{nstance} segmentation is significantly more challenging than semantic segmentation because it requires identifying every individual instance of a class, and the visible extent of each. The information recovered has proven invaluable for scene understanding, however.
With the easy availability of various 3D sensors, such as LiDAR and laser scanner, 3D understanding is drawing increasing attention for its numerous promising applications, such as autonomous driving \cite{shi2019pointrcnn, shi2020part, shi2020pv, yang2019std, Shi_2020_CVPR}, virtual reality \cite{Li_cvpr_frodo, lin2019photometric, sun_pix3d, Mescheder2019CVPR, Peng2020ECCV}, to name a few. 
Although significant advances have been made in the 2D image domain \cite{he2018maskrcnn, chen2020cascade, chen2020blendmask, tian2020conditional}, the counterpart in 3D point cloud has proven far more challenging, due to the inherent irregularity, sparsity of the data, and larger searching space of 6 Degrees-of-Freedom (DoF). Directly extending the methods from 2D image to 3D point cloud might lead to inferior performance.
By way of example, Mask R-CNN \cite{he2018maskrcnn}, which has shown great success in the 2D domain, performs poorly when directly utilized in the 3D applications \cite{hou20193dsis}. 

Existing methods have explored several ways to address this challenging task. Most top-performing approaches apply a bottom-up strategy. PointGroup \cite{jiang2020pointgroup} generates proposals of instances by gradually merging neighboring points that have the same category label. Both original and centroid-shifted points are explored with a manually specified searching radius, followed by a ScoreNet to estimate the objectness of the proposals. Some other approaches \cite{Zhang_2021_CVPR, he2020eccvembedding, he2020eccvmemory, Engelmann20CVPR, Jean2019mtml, wang2019asis, wang2018sgpn, Liu2019masc, han2020occuseg} predict point-wise (or patch-wise) embedding to group points (or patches) into individual instances. For testing, an extra clustering algorithm, such as mean-shift or greedy clustering, is required to obtain the final results. Although promising performance has been achieved on benchmarks like ScanNetV2 \cite{dai2017scannet} and S3DIS \cite{armeni2016s3dis}, these bottom-up methods often suffer from several drawbacks:

\begin{itemize}
	\item The performance of these methods is sensitive to the values of the heuristic hyper-parameters. For example, the predefined neighbouring searching radius in PointGroup \cite{jiang2020pointgroup} and 3D-MPA \cite{Engelmann20CVPR} has a great impact on the final accuracy, as the value chosen directly determines the quality of the instance proposals. Modifying the clustering radius from 3cm to 2cm results in mAP of PointGroup dropping by more than 6\%, and more than 20\% if the value is set to 1cm. Qualitative results are shown in Figure \ref{fig:main}, illustrating the method's limited robustness and generalization capability. In addition, a large number of empirical settings also need to be carefully tuned, for example, the embedding dimension, loss weights, distance thresholds for positive and negative examples, and the value of bandwidth in mean-shift clustering. In \cite{Engelmann20CVPR}, an additional binary segmentation subnetwork is required to refine the foreground points for each cluster.
	
	\item Embedding-based methods encourage the learned embeddings of the same instance to be pulled to each other while pushing those of different instances as far away from each other as possible. The key to the success is to learn discriminative embeddings. To this end, \cite{he2020eccvembedding, he2020eccvmemory, wang2019asis} designed complicated loss functions. \cite{Jean2019mtml} proposed to optimize a Multi-Value Conditional Random Fields (MV-CRF). \cite{han2020occuseg} requires a pre-processing step, graph cut clustering, to obtain a bunch of super-voxels.  These methods ignore the spatial distributions of instances and introduce complex pre- and post-processing steps, rendering them unsuitable for real-time applications such as robotics and virtual reality.
 
	
\end{itemize}

There also exists a few top-down based methods \cite{yang20193dbonet, hou20193dsis, yi2018gspn}, which first detect 3D bounding boxes, and then instance segmentation is obtained by finding out the foreground points. However,  due to the large variations of sizes, 3D detection for indoor scenes remains an open problem and still far from satisfactory. Building a strong dependency upon detections often leads to sub-par performance and joint/fragmented instances. 

\revise{For the task of 3D instance segmentation, two types of knowledge are required: content (or appearance) and location. In contrast to traditional CNNs/MLPs, which struggle at achieving these due to the shareable weights, we propose a novel framework for 3D instance segmentation, \SexyName, which surpasses previous top-performing approaches with strong robustness by applying dynamic convolution. }
Inspired by the works of \cite{tian2020conditional, debrabandere16dynamic}, \SexyName encodes instances as a continuous function, in which the parameters are generated on the fly in response to the characteristics of instances. 
Different from 2D images, point cloud from 3D manifold faces a major challenge: data are collected from the surface of objects and may far from the object centroids, causing failure to directly use the methods of \cite{tian2020conditional, tian2019fcos, tian2021fcos} which utilize the predictions from the central areas of instances to achieve higher accuracy. 
To address the issue, we follow \cite{qi2019deep} by first predicting the centroids of instances. A grouping algorithm \cite{jiang2020pointgroup} is then applied to explore the homogenous points that have close votes for instance centroids and identical semantic labels. 
A sub-network is deployed to explore the spatial distribution of each cluster and output the parameters to address the characteristics of instances. The parameters are expected to contain both geometric and semantic knowledge of the instances.
Specifically, unlike previous top-down methods that explicitly encode position information by predicting 3D bounding boxes, our method, on the other hand, implicitly encodes it by concatenating relative position to the mask head. In addition, to further encode the category of the instance, we implement it in a straightforward way by constraining the convolution to the points that have the same category with the cluster. For example, a specific chair is masked out from the points that are classified as `Chair', instead of all points of the scene.
For masks generation, voxels(points) belonging to the specific instance will be segmented out through several simple convolutional layers. 
\revise{}

\begin{figure}[bt]
	\centering
	{
		\includegraphics[width=8.6cm]{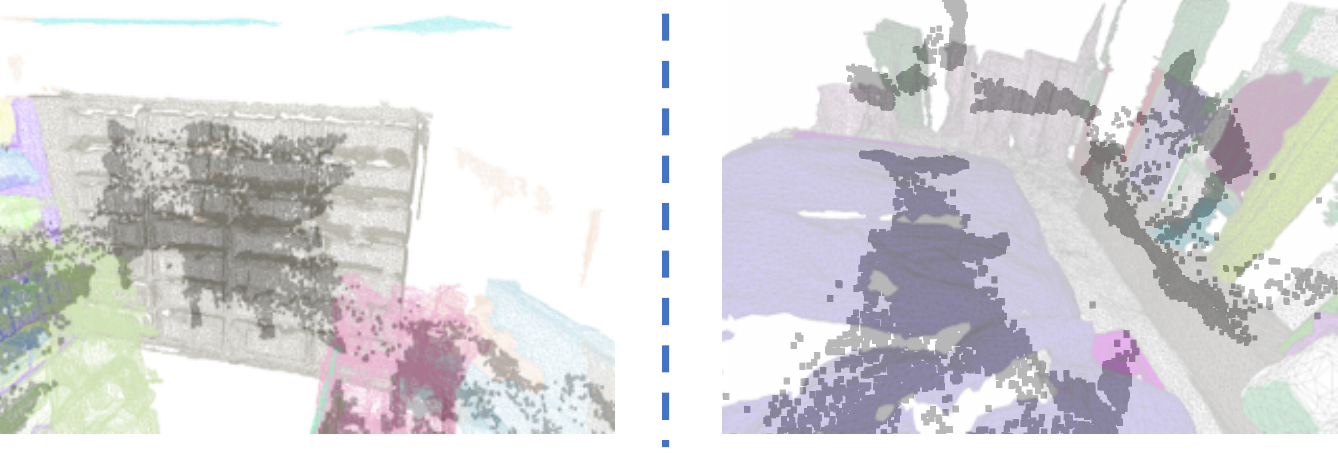}
	}
	\caption{The illustration of per-point centroids prediction by current sparse convolution backbones \cite{graham2018sparseconv, choy20194d}. Different instances are presented with random colors. Points with gray color are the predicted centroids. It is difficult for the points that are far from the centroids due to the limited context information. Directly depending on the grouping algorithm \cite{jiang2020pointgroup, Engelmann20CVPR} for generating instance proposals often leads to sub-par performance and joint/fragmented instances.
	}
	\label{fig:cen_pred}
\end{figure}

Besides, as has been proved in the 2D image domain \cite{cheng2020panoptic, deeplabv3plus2018, ross_2014_cvpr_rich, wang_2015_torward,  dai_2017_deformable}, rich context information is critical to the success of the perception tasks. We observe that in modern top-performing backbones \cite{graham2018sparseconv, choy20194d} for 3D instance segmentation, context is often limited, as illustrated in Figure \ref{fig:cen_pred}. Unlike objects of small sizes, such as `Chair' and `toilet' that have precise centroids predictions, points that are far from the centroids in large-size objects perform poorly due to the limited context. 
We address the problem by introducing a small transformer module \cite{Vaswani2017attention} to capture a long-range dependency and build high-level interactions among different regions. As the module is built upon the bottleneck layer, only limited computational overheads are introduced. 

Our primary contributions are summarized as follows. (1) We propose \SexyName, a novel framework for 3D point cloud instance segmentation, which is free from box prediction and embedding learning by using dynamic convolutions. \SexyName is tailored for 3D point cloud, in which instances are encoded in a continuous function, conditioned on both relative position and categories. 
(2) Our proposed method presents strong robustness to the manually set hyper-parameters. Both qualitative and quantitative results are provided. 
(3) \SexyName shows superiority in both effectiveness and efficiency. We conduct comprehensive experiments on various datasets, including both large-scale indoor scenes and CAD scanned objects. To further validate the effectiveness, we implement our method with both voxel-based and point-based architectures. The consistent improvements indicating that \SexyName can serve as a strong alternative for 3D instance-level recognition tasks.


	\section{Related Work}
\textbf{Feature Extraction on 3D Point Cloud.} In contrast to the image domain, wherein the feature extraction is relatively mature (see \eg VGGNet \cite{Simonyan15}, ResNet \cite{He2016resnet} and ViT \cite{dosovitskiy2020image}), methods for 3D point cloud representation are still developing. The most prevalent approaches can be roughly categorised as: point-based \cite{qi2017pointnet, qi2017pointnetplusplus, Zhao2021_pointtransformer, hu2020randla, Wu_2019_CVPR, xu2018spidercnn, thomas2019KPConv}, voxel-based \cite{maturana2015voxnet, wu20153dshapenet, graham2018sparseconv, choy20194d, Riegler2017OctNet, song2016ssc}, projection-based \cite{su2015multiview, qi2016multiview, dai20183dmv, chen_2017_multiview, kanezaki2018_rotationnet, Lang_2019_CVPR, Tat2018} and hybrid methods \cite{lang2019pointpillar, liu2019pvcnn, shi2021pv, shi2020pv}. 

PointNet \cite{qi2017pointnet} is one of the pioneering point-based deep learning approaches. It exploits a shareable multi-layer perceptron (MLP) network to extract per-point representation. PointNet++ \cite{qi2017pointnetplusplus} extends this approach by introducing a nested hierarchical PointNet to extract local context information.
Some methods \cite{thomas2019KPConv, Wu_2019_CVPR, xu2018spidercnn} enhance the representation capability by using a continuous convolution to construct dynamic weights for each input point. Although simple, point-based methods have been widely applied in the tasks of 3D object detection \cite{qi2019deep, qi2020imvotenet, Yang_2020_CVPR, yang20193dbonet}, semantic segmentation \cite{xu2018spidercnn, thomas2019KPConv, hu2020randla}, and instance segmentation \cite{wang2018sgpn, wang2019asis, he2020eccvembedding, he2020eccvmemory, yi2018gspn}.

An alternative approach is transforming unordered points to regular 3D voxels. 3D convolution is then applied. The traditional 3D convolution operation is constrained by inefficient computation and limited GPU memory, as most resources are wasted on void space. In contrast, sparse convolution \cite{choy20194d, graham2018sparseconv, tang2020searching} is proposed to address the problem by focusing the computation on the data, rather than the space it occupies. Some other methods \cite{Riegler2017OctNet, Lei_2019_octree, matheus_2018_eccv_tree, Shu_2019_ICCV} address the sparsity of the data by partitioning the 3D space into a set of structured trees. 

Considering the success of convolution neural networks in the 2D image, a number of methods are proposed to project 3D point cloud to multi-view representation. 2D CNN is used to extract features, followed by a multi-view fusion module. 3DSIS \cite{hou20193dsis} proposes to jointly fuse multi-view 2D RGB features with 3D geometric features. \cite{Tat2018} operates directly on surface geometry by projecting local structure on the tangent plane for every point. \cite{hu2021bpnet, Gaddetpami, Liangcontfuse, su18splatnet} achieves multi-modality fusion by merging both 2D and 3D features for perception tasks.

\begin{figure*}[htbp]
	\begin{center}
		{
			\includegraphics[width=17.5cm]{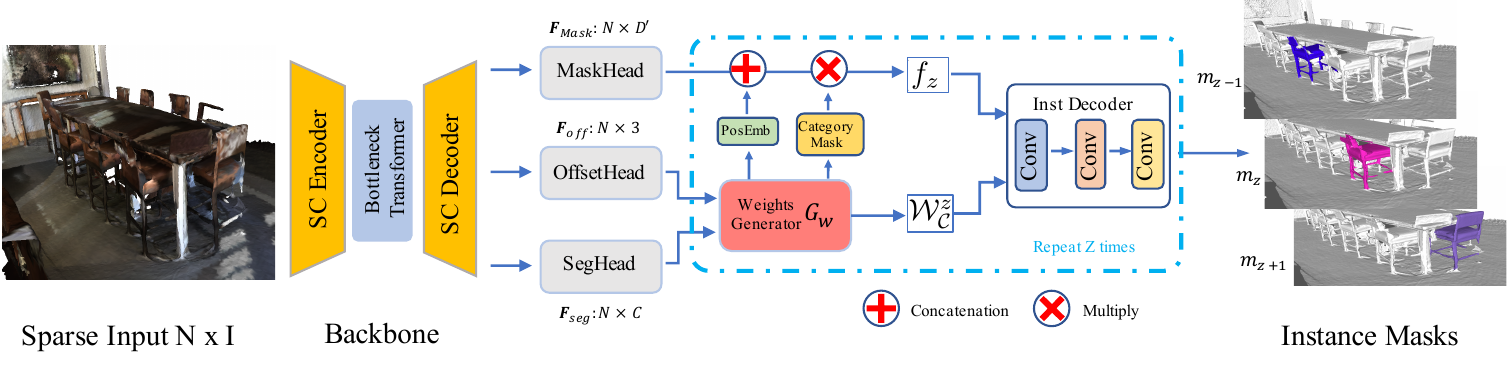}
			
		}
	\end{center}
	\caption{
		The structure of \SexyName. It contains three main components: (1) a sparse convolution backbone based on~\cite{graham2018sparseconv}, which contains a light-weight transformer and outputs three parallel heads: mask head, offset head, and semantic segmentation head.  (2) A weight generator that takes both centroid predictions and semantic segmentation as input. Homogenous points that have close votes for instance centroids and share the category predictions are explored to output instance-aware position embeddings, category-specific masks, and convolutional filters. (3) An instance decoder. Binary masks of instances are decoded by applying several convolutions, with the filters constructed by the Weight Generator. The second part and the third part are repeated in parallel for $Z$ times, where $Z$ is the total number of the instance candidates. 
	}
	\label{fig:framework}
\end{figure*}

\textbf{Instance Segmentation of 3D Point Cloud.} Similar to the 2D image domain, instance segmentation on the 3D point cloud can be roughly categorized into two groups: top-down and bottom-up. Top-down methods often follow a detect-then-segment pipeline, which first detects 3D bounding boxes of the instances and then predicts the foreground points within the boxes. 3D-BoNet \cite{yang20193dbonet}, for instance, first detects unoriented 3D bounding boxes from a single global representation by utilizing a Hungarian matching algorithm. For each bounding box, foreground points are selected by using point-level features. GSPN \cite{yi2018gspn} leverages an analysis-by-synthesis strategy, in which object proposals are generated by first reconstructing them. Compared to the top-down approaches, wherein the 3D detection is often suboptimal, bottom-up approaches are more straightforward and have achieved promising performance. They group sub-components into instances. Methods following this pipeline have dominated the leaderboard of the ScanNet dataset \cite{dai2017scannet}\footnote{\url{http://kaldir.vc.in.tum.de/scannet_benchmark/}}. The grouping techniques include point-wise metric learning \cite{wang2019asis, he2020eccvmemory, phamjsis3dcvpr19, he2020eccvembedding, Zhang_2021_CVPR}, patch-wise metric learning \cite{han2020occuseg, Engelmann20CVPR}, and spatial aggregation \cite{jiang2020pointgroup, Engelmann20CVPR}. ASIS \cite{wang2019asis}, for example, learns point-level embeddings and is supervised by a discriminative loss function \cite{bra2017cvprdis}, which encourages points belonging to the same instance to be mapped to close locations in a metric space while separating points belonging to different instances. Mean-shift algorithm is then applied to group individual points. 
OccuSeg \cite{han2020occuseg} first divides points into super-voxels based on the clues of both appearance and position information by using a graph-based segmentation algorithm \cite{Pedro2014_ijcv}. Patch-wise embedding is learned to merge patches that belong to the same instance. 
3D-MPA \cite{Engelmann20CVPR} proposes to cluster patches through a Graph Neural Network. 
Some other works, \eg PointGroup \cite{jiang2020pointgroup}, generate proposals by exploring the void space between objects. Both original and centroid shifted sets are used to group points. \revise{ SoftGroup~\cite{vu2022softgroup} extends the idea by considering the soft labels of semantic segmentation for aggregating points. Each point can be associated with multiple categories to address the misclassification from semantic segmentation. SSTNet~\cite{liang2021instance} introduces a semantic superpoint tree to merge intermediate nodes in a bottom-up manner. HAIS~\cite{Chen_HAIS_2021_ICCV} proposes a hierarchical aggregation strategy by considering both point-level and set-level clusters.}

\textbf{Dynamic Convolution.} Our approach is related to the recent works of \cite{tian2020conditional, debrabandere16dynamic, Chen_2020_CVPR_dynamic, wu2018pay}, wherein the dynamic convolutions are built upon light-weight modules to predict kernels for convolution. The major difference with regular convolution is the weights are not shared across different positions but rather generated dynamically at every position. 
\cite{Chen_2020_CVPR_dynamic} proposed to increase the model complexity by assembling multiple kernels in parallel. Attention weights are generated via an input-dependent linear projection layer. 
\cite{Wu_2019_CVPR} came up with Lightconv by predicting different convolution kernels that are depth-wise separable. 
AdaptIS \cite{Konstantin_adaptis2019} adapts to the input point and produces different binary masks by predicting the parameters for batch normalization.
CondInst \cite{tian2020conditional} attempts to address the 2D instance segmentation by applying a dynamic mask head. Information about an instance is encoded in the convolutional parameters. As the method is embedded with a detection pipeline FCOS \cite{tian2019fcos}, only the central areas of an instance are involved in mask decoding, which releases the burden for heavy computation and is free from optimization difficulties. 
However, due to the sparse nature of the 3D point cloud data, the points collected from the surface of objects may be far from the corresponding geometric centroids, making it hard to achieve selective predictions as in \cite{tian2020conditional, tian2021fcos}. In addition, the complexity of the scene in the 3D domain also makes it hard to discriminate the instances of different categories. In contrast, \SexyName decodes masks of instances conditioned on both geometric and category predictions. As shown in the experiments, \SexyName can achieve much better performance than CondInst and surpass previous methods on multiple benchmarks in both effectiveness and efficiency.
	
	\section{Proposed Methods: \SexyName for Robust 3D Instance Segmentation.}
In \SexyName, the core idea is to produce $Z$ sets of data-dependent convolutional parameters and parallel instance decoders in response to the characteristics of $Z$ different instances. Each decoder contains a set of essential representations of a specific instance, which will be used to distinguish it from all instances. Both category and position information are incorporated in the decoding process. Besides, as surface points are usually far from the geometric centroid, it is hard to constrain the prediction within the central areas, as is widely used in the 2D image domain \cite{tian2019fcos, tian2020conditional, tian2021fcos, zhou2019objects} to (1) provide accurate predictions and (2) reduce the computation cost in testing. 
To address this, a simple clustering algorithm is applied to explore the spatial distribution of each instance and incorporate a large context. 
At last, although sparse convolution neural networks \cite{graham2018sparseconv, choy20194d} have been proposed for efficient computation, most of the current models still limited to light-weight architectures, which are comparable to the ResNet18 and ResNet34 that are used in the 2D image. We propose to apply a Transformer module \cite{Vaswani2017attention} on the bottleneck layer to incorporate a large context with negligible computation.

\subsection{Overall Architecture}
The structure of \SexyName is depicted in Figure~\ref{fig:framework}. 
The input is a 2D matrix recording the point features $\textbf{P} \in \mathbb{R}^{N\times I}$, where $N$ is the total number of points and $I$ is the dimension of each point feature.
The goal is to predict a set of point-level binary masks and their corresponding category labels, denoted as $\{(\hat m_k, \hat c_k)\}$, where $\hat m_k \in \{0,1\}^N$, and $\hat c_k \in \{1, 2, \cdots, C\}$. $C$ denotes the category number.
Compared with previous top-performing approaches \cite{jiang2020pointgroup, Engelmann20CVPR}, where masks of instances are dependent on the proposals, 
our method is proposal-free and can produce instance masks using only a small number of $1\times1$ convolutional layers. 
The associated convolution filters are dynamically generated, conditioned on both spatial distribution of the data and the semantic predictions. As shown in Figure~\ref{fig:framework}, \SexyName is comprised of three primary components:
(1) A backbone network which can be either volumetric \cite{graham2018sparseconv, choy20194d} or point-based \cite{qi2017pointnet, qi2017pointnetplusplus}. A light-weight transformer module is introduced, aiming to incorporate a large context and capture long-range dependencies. 
(2) A weight generator, that is achieved by grouping and re-voxelizing homogeneous points, is proposed to respond to the individual characteristics of each instance.
(3) An instance decoder. Binary masks of instances are obtained in parallel, by applying several simple convolution layers with kernels of 1$\times$1. Both position and category dependencies are added during the decoding. 

\subsection{Backbone Network}
\label{section:backbone}
Our method is not restricted to any specific choice of 3D point cloud representation, both sparse convolution \cite{graham2018sparseconv, choy20194d} and point-based backbones \cite{qi2017pointnetplusplus} can be used. Following \cite{graham2018sparseconv, jiang2020pointgroup}, we construct a U-Net, which consists of an encoder and a decoder that have symmetrical structures. 
However, recent architectures are often constrained by a limited receptive field and representation capability, due to the small number of convolution layers and channels.
We demonstrate this by regressing the centroids of instances.
As shown in Figure~\ref{fig:cen_pred}, points that are far from the centroid of an instance often show limited capability to predict the accurate offsets, leading to inaccurate grouping clusters. 
Recently, transformer modules have drawn much attention in various vision tasks \cite{dosovitskiy2020, Nicolas2020detr}, partly due to its capability to capture long range dependencies. However, extra computational overheads are required due to the heavy quadratic calculations in self-attention.
To this end, we propose a light-weight transformer \cite{Vaswani2017attention} on top of the encoder, which contains a much shorter length of tokens. 
The transformer module only contains the feature encoding part with only two multi-head self-attention layers, each of which is followed by a feed-forward layer.  We apply relative position encoding for the feature learning. As only a small number of points in the bottleneck layer, very limited computing is added.

The output of the backbone is denoted as $\textbf{F}_b \in \mathbb{R}^{N \times D}$, where $D$ is the dimension of the output channel. Three parallel heads are built upon $\textbf{F}_b$ for semantic segmentation ($\textbf{F}_\text{seg} \in \mathbb{R} ^{N \times C}$), offset prediction ($\textbf{O}_\text{off} \in \mathbb{R}^{N \times 3}$), and instance masking ($\textbf{F}_\text{mask} \in \mathbb{R}^{N\times D'}$), respectively, where $C$ is the category number ($C$ is 20 for ScanNetV2 and 13 for S3DIS). 

\textbf{Semantic Segmentation Head.} We apply traditional cross entropy loss $\mathcal{L}_\text{seg}$ for semantic segmentation. Pointwise prediction of the category label can be easily obtained, indicated as $\{l_\text{seg}^i\}_\text{i=1}^N$.

\textbf{Centroid Offset Head.} 
One of the inherent challenges in instance segmentation from point clouds is the density with which an object is sampled depends on the sensor distance and occlusions. In addition, the instances from indoor scenes often have a large variations in both sizes and aspect ratios. Thus, using a unified radius for searching often causes the vulnerability of model.
To address the problem, we follow VoteNet \cite{qi2019deep}, by shifting points towards the corresponding centroids of instances. Point-wise prediction $o_\text{off}^i$ is supervised by the following loss function:
\begin{equation}
\mathcal{L}_\text{ctr} = \frac{1}{N_v}\sum_{i=0}^{N}\| p^i + o_\text{off}^i- ctr_\text{gt}^i\|\cdot \mathds{1}(p^i)
\end{equation} 
where $p^i$ is the coordinates of the $i$-th point, $o_\text{off}^i$ is the offset prediction of the $i$-th point, and $ctr_\text{gt}^i$ is the geometric centroid of the corresponding instance. $\mathds{1}(p^i)$ is an indicator function, representing whether $p^i$ is a valid point for centroid prediction. $N_v$ is the total number of the valid points. For example, the categories of `floor' and `wall' are ignored for instance segmentation in ScanNetV2 \cite{dai2017scannet}, making them free from offset predictions. 

\textbf{Masking Head.} 
The masking head predicts a set of features ($\textbf{F}_\text{mask} \in \mathbb{R}^{N\times D'}$) through an MLP layer, where $D'$ is the number of the output channel. It is shared for all the instance decoders and taken as inputs to predict the desired instance mask in a fully convolutional fashion.

\begin{figure*}[htbp]
	\centering
	{
		\includegraphics[width=16cm]{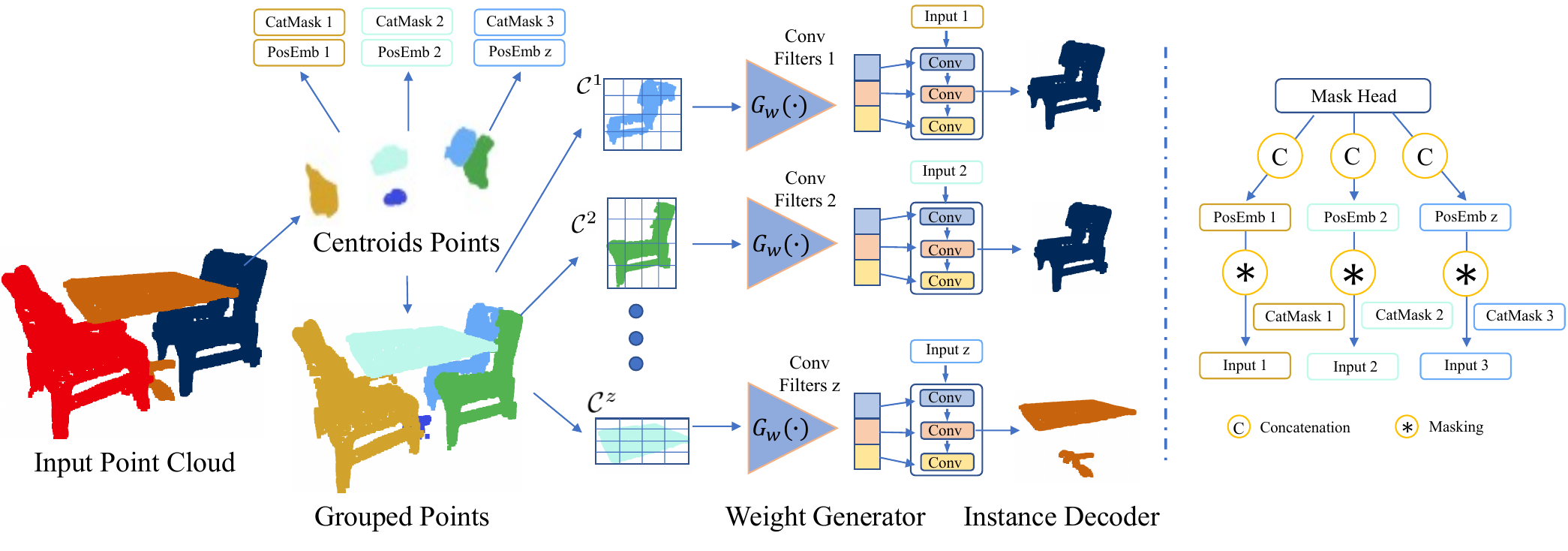}
	}
	\caption{ The illustration of the weight generator and instance decoder. Homogenous points are clustered by exploring both category prediction and geometric distribution. A light-weight sub-network is then applied to incorporate the larger context and applied once for each cluster to generate the convolution parameters used in instance decoding. Each filter is responsible for one instance. Except the filters, the input features for the instance decoding incorporates both category and relative position knowledge.
	}
	\label{fig:generator}
\end{figure*}

\subsection{Dynamic Weight Generator}
\label{section: filter}
Due to the larger searching space of 6 Degrees-of-Freedom and heavy computational overheads, recent network architectures \cite{graham2018sparseconv, choy20194d} often select backbones with small channels and layer scalability, resulting in limited representation capability for point cloud data. To this end, we propose to first group homogenous points that have close votes for the geometric centroids and share the category predictions. Then instance-aware filters are dynamically generated by applying a small sub-network for a large context aggregation, as shown in Figure \ref{fig:generator}.
Provided both predicted semantic labels and centroids offsets, we are ready for grouping homogenous points by using a similar strategy that is used in \cite{jiang2020pointgroup}. Details of the algorithm can be found in Algorithm \ref{alg: cluster}. However, different from \cite{jiang2020pointgroup} that directly treats the clusters as individual instance proposals, our method explores the spatial distribution of these points and generates filters in response to the characteristics of instances.
Thus, our proposed method removes the dependencies between the quality of the proposals and the final performance. Comprehensive experiments demonstrate that our method is robust to the pre-defined hyper-parameters. Qualitative results can be seen in Figures \ref{fig:main}. The closest work to our approach is CondInst \cite{tian2020conditional}, where instance-aware parameters are generated for each valid pixel in the central area. However, we demonstrate that CondInst is not well suited for 3D point cloud for the following reasons: (1) As points are collected in the surface of objects, central areas are hard to define, leading to a heavy burden for computation and difficulties for optimization in training. (2) The filters are responsible for discriminating one instance from all other points, failing to adopt category knowledge. To address the above problems, we propose to : (1) generate filters for each homogenous cluster instead of each point. (2) restrict each filter responsible for separating instances from only one category.

\begin{algorithm}[t]
	
	\caption{Clustering algorithm.
		$N$ is the number of input points.
		$Z$ is the total number of the clusters.}
	\label{alg: cluster}
	\hspace*{0.02in} {\bf Input:} pre-defined clustering radius $r$; \\
	\hspace*{0.44in}
	coordinates $\{p^1, p^2, ..., p^N\}\in \mathbb{R}^{N \times 3}$. \\
	\hspace*{0.44in}
	semantic predictions $\{ l_\text{seg} ^1, l_\text{seg}^2, ..., l_\text{seg}^N \} \in \mathbb{R} ^ N$. \\
	\hspace*{0.44in}
	centroid predictions $\{ o_\text{off}^1, o_\text{off}^2, ..., o_\text{off} ^ N\} \in \mathbb{R}^{N\times 3}$. \\
	\hspace*{0.02in} {\bf Output:} clusters $\{\mathcal{C}^1, \mathcal{C}^2, ..., \mathcal{C}^Z\}$. \\
	\hspace*{0.44in} 
	cluster centroids $\{ \mathcal{C}_\text{ctr}^1, \mathcal{C}_\text{ctr}^2, ..., \mathcal{C}_\text{ctr}^Z\}$. \\
	\hspace*{0.44in}
	cluster semantic labels $\{ l_\mathcal{C}^1, l_\mathcal{C}^2, ..., l_\mathcal{C}^Z \}$.
	\begin{algorithmic}[1]
		\State initialize an empty cluster set, $\mathcal{C}$.
		\State initialize an empty centroids set of the clusters, $\mathcal{C}_\text{ctr}$.
		\State initialize an empty semantic labels set of the clusters, $l_\mathcal{C}$.
		\State initialize an array $v$ of length $N$ with all zeros for recording the visiting status. 
		\For{$i = 1$ to $N$}
		\If{$l_\text{seg}^i$ is a stuff class (\eg, floor)}
		\State $v_i$ = True
		\EndIf
		\EndFor
		
		\For{$i = 1$ to $ N$}
		\If{$v_i$ == False}
		\State initialize an empty queue $Q$
		\State initialize an empty cluster $\text{C}$
		\State initialize the point number for $N_\text{C}$ to be zero
		\State $v_i$ = True; $Q$.enqueue($p^i$); 
		$\text{C}$.add($p^i$); 
		
		\While{$Q$ is not empty}
		\State $p^k = Q$.dequeue()
		\For{$j \in [1, N]$}
		\If{ $v_j$==True}
		\State continue
		\EndIf
		\If{$||p^j+o_\text{off}^j - p^k-o_\text{off}^k||_2 < r$ and \\ \hspace*{1.0in}$l_\text{seg}^j == l_\text{seg}^k$}
		\State $v_j $=True; $Q$.enqueue($p^j$); $\text{C}$.add($p^j$)
		\State $N_\text{C}$ += 1
		\EndIf
		\EndFor
		\EndWhile
		
		\State $\mathcal{C}$.add($\text{C}$)

		\State $\mathcal{C}_\text{ctr}$.add($\frac{1}{N_C}\sum\text{C}$)
		
		\State $l_\mathcal{C}$.add($l_\text{seg}^i$)

		\EndIf
		\EndFor
		\State \Return $\mathcal{C}$, $\mathcal{C}_\text{ctr}$, $l_\mathcal{C}$
	\end{algorithmic}
	
\end{algorithm}

Provided the offset predictions $\{o_\text{off}^i \} _\text{i=1}^N$, distribution of centroids $\{p_\text{ctr}^i \in \mathbb{R}^3 \}_\text{i=1}^N$ can be easily calculated: $p_\text{ctr}^i=p^i + o_\text{off}^i$. With $\{p_\text{ctr}^i\}_{i=1}^N$ and semantic labels $\{l_\text{seg}^i \} _{i=1} ^N$, instances of different categories are separated to a certain extent. We explore the void spaces among instances by applying a breadth-first searching algorithm (see Algorithm \ref{alg: cluster} for details) to group homogenous points that have identical semantic labels and close centroids predictions.
Point $p^j$ can be grouped with $p^i$ if it satisfies:
(1) $l_\text{seg}^j = l_\text{seg}^i$. (2) $\|p_\text{ctr}^j - p_\text{ctr}^i\|_2 <= r$, where $r$ is a pre-defined searching radius. The grouping algorithm ends up with a set of clusters $\{\mathcal{C}^z\}_\text{z=1}^Z$, where $Z$ refers to the total number of clusters. 
As only one specific category is considered for each cluster, the semantic label $l_\mathcal{C}^z \in \mathbb{R}$ of cluster $\mathcal{C}^z$ can be easily obtained. We define $l_\mathcal{C}^z = c$ if $c$ is the dominant category of the points in cluster $\mathcal{C}^z$.
We also label the geometrical centroids for cluster $\mathcal{C}^z$ as $\mathcal{C}_\text{ctr}^z \in \mathbb{R} ^3$, which is calculated as the average of the coordinates of the points in $\mathcal{C}^z$.
Thus, each cluster contains a bunch of points that are distributed across the whole instance or part of the instance, introducing a larger context and rich geometric information than a single point. 

For cluster $\mathcal{C}^z$, we first voxelize it with a grid size of $g$, which is set to 14 in all our experiments. The feature of each grid is calculated as the average of the point feature $\textbf{F}_b$ within the grid, where $\textbf{F}_b$ is the output of the backbone. 
To aggregate the context for cluster $\mathcal{C}^z$, a light-weight sub-network $G_w(\cdot)$ is proposed. It contains two sparse convolutional layers with a kernel size of 3, a global pooling layer, and an MLP layer. The output is all convolutional parameters flattened in a compact vector, $\mathcal{W}_\mathcal{C}^z$. Each $\mathcal{W}_\mathcal{C}^z$ is responsible for one specific instance.
The size of $\mathcal{W}_\mathcal{C}^z$ is determined by the feature dimension and the number of the subsequent convolution layers (see Eq. \ref{eq:filter_calcu}).

\subsection{Instance Decoder}
\label{section: instance_gen}
Given a specific category, position representation is critical to separate different instances. To let the generated parameters be aware of the positions, we directly append position embeddings in the feature space. For the $z$-th instance with geometric centroid of $\mathcal{C}_\text{ctr}^z$, position embedding for the $i$-th point $f_\text{pos}^i$ is calculated as:
\begin{equation}
f_\text{pos}^i = p^i - \mathcal{C}_\text{ctr}^z
\end{equation}
where $p^i$ is the coordinates of the $i$-th point. For $\mathcal{W}_\mathcal{C}^z$, the input feature $\{f_z^i \in \mathbb{R}^{D'+3} \}_{i=1}^{N} $ is produced by concatenating $\{f_\text{pos}^i \in \mathbb{R}^3 \}_{i=1}^N$ and $\textbf{F}_\text{mask} \in \mathbb{R}^{N\times D'}$

Provided both instance-aware filters $\{\mathcal{W}_\mathcal{C}^z\}_{z=1}^Z$ and position-embedded features $\{f_z \in \mathbb{R}^{N\times (D'+3)} \}_{z=1}^Z$, we are ready to decode binary segmentations of instances. The whole decoder contains several convolution layers with a kernel size of $1\times1$. Each layer uses ReLU as the activation function without normalization. Supposing the feature dimension of $f_\text{mask}^i$ is 8, meaning $D'=8$, the number of convolution layers are 3, and the feature dimension of the decoder is 8, the total number of parameters (including both weights and biases) of $\mathcal{W}_\mathcal{C}^z$ is 177, which is calculated by:

\begin{equation}
177 = \underbrace {(8+3) \times 8 + 8}_{conv1} + \underbrace{8 \times 8 + 8}_{conv2} + \underbrace{ 8 \times 1 + 1}_{conv3}
\label{eq:filter_calcu}
\end{equation}

Formally, the instance decoder is formulated as:
\begin{equation}
m_z = Conv(\mathcal{W}_\mathcal{C}^z, f_z) \times b_z
\end{equation}
where $m_z \in \mathbb{R}^N$ is the predicted segmentation for the $z$-th instance and $b_z \in \mathbb{R}^N$ is a binary mask, indicating whether the points have identical semantic predictions with $l_\mathcal{C}^z$. 

During training, the supervision for $\mathcal{C}^z$ is $z$-th binary ground truth $\hat m_z$ if $\mathcal{C}^z$ has the largest portion of $z$-th instance. 
The loss function for instance segmentation is defined as:
\begin{equation}
\mathcal{L}_\text{mask} = \frac{1}{Z} \sum_{z=1}^{Z} \frac{1}{N_z} \sum_{j=1}^{N} \mathds{1}(l_\text{seg}^j==l_\mathcal{C}^z) \cdot L_\text{BCE}(m_z^j, \hat m_z^j)
\label{eq:conv}
\end{equation}
where $Z$ is the total number of the clusters, $l_\text{seg}^j$ is the semantic prediction of the $j$-th point, $l_\mathcal{C}^z$ is the semantic label of the $z$-th cluster, and $L_\text{BCE}$ is the binary cross entropy loss function.
$\mathds{1}$ is an indicator function, showing the loss is only computed on the points that have identical semantic labels with group $\mathcal{C}^z$, and $N_z$ is a normalization item which is calculated as: $\sum_{j=1}^N \mathds{1}_{l_\text{seg}^j=l_\mathcal{C}^z}$.
In addition to the point-wise supervision, we also utilize the dice loss \cite{tian2020conditional} $\mathcal{L}_\text{dice}$, which is designed for maximizing the intersection-over-union (iou) between the prediction and the ground truth. 

\subsection{Implementation details}
The loss function of \SexyName contains four parts and can be formulated as:
\begin{equation}
\mathcal{L} = \mathcal{L}_\text{seg} + \mathcal{L}_\text{ctr} +  \mathcal{L}_\text{mask} + \mathcal{L}_\text{dice}
\end{equation}
where $\mathcal{L}_\text{seg}$ is for semantic segmentation, 
$\mathcal{L}_\text{ctr}$ is for instance centroids supervision, and
$\mathcal{L}_\text{mask}$ and $\mathcal{L}_\text{dice}$ are two loss items for instance segmentation. All loss weights are set to 1.0. 

To demonstrate the effectiveness of our proposed method, we implement \SexyName with volumetric \cite{choy20194d, graham2018sparseconv} architecture and extend \SexyName to a point-based \cite{qi2017pointnetplusplus} pipeline. 

For the volumetric architecture, we evaluate the performance on two large scale indoor datasets: S3DIS \cite{armeni2016s3dis} and ScanNetV2 \cite{dai2017scannet}, which contains rich annotations for instance segmentation. The voxel size is set to 0.02m and 0.05m for ScanNetV2 \cite{dai2017scannet} and S3DIS \cite{armeni2016s3dis}, respectively. 
We utilize a training strategy with a batch size of 16 across 4 GPUs. For the first 12k iterations, we only train the semantic segmentation $\mathcal{L}_\text{seg}$ and centroid prediction $\mathcal{L}_\text{ctr}$. For the next 38k iterations, we compute all the loss items. The initial learning rate is set to 0.01 with an Adam optimizer. We apply the same data augmentation strategy with \cite{jiang2020pointgroup}, including random cropping, flipping, and rotating. For inference, we perform non-maximum-suppression (NMS) on the instance binary masks $\{m_\mathcal{C}^z\}_{z=1}^Z$. Although introducing an extra scorenet \cite{jiang2020pointgroup} for predicting the objectness of each instance proposal is feasible, extra computational overheads are required. We propose to use the mean value of semantic scores among the foreground points as the objectness scores for instances.
The IoU threshold is the same as \cite{jiang2020pointgroup}, with a value of 0.3. Cluster $\mathcal{C}^z$ is ignored if it contains points less than 50.

Also, we extend the idea of dynamic convolution to a point-based architecture for instance segmentation and evaluate the performance on the PartNet \cite{mo_2019_partnet} dataset.

\section{Experiments}

We evaluate \SexyName on various datasets, including large-scale indoor scans, \eg ScanNetV2 \cite{dai2017scannet} and Stanford 3D Indoor Semantic Dataset (S3DIS) \cite{armeni2016s3dis}, and fine-grained 3D objects benchmark: PartNet \cite{mo_2019_partnet}. 
The performance on both volumetric and point-based architectures are reported to illustrate the effectiveness of our proposed method. 


\begin{figure*}
	\centering
	{
		\includegraphics[width=17cm]{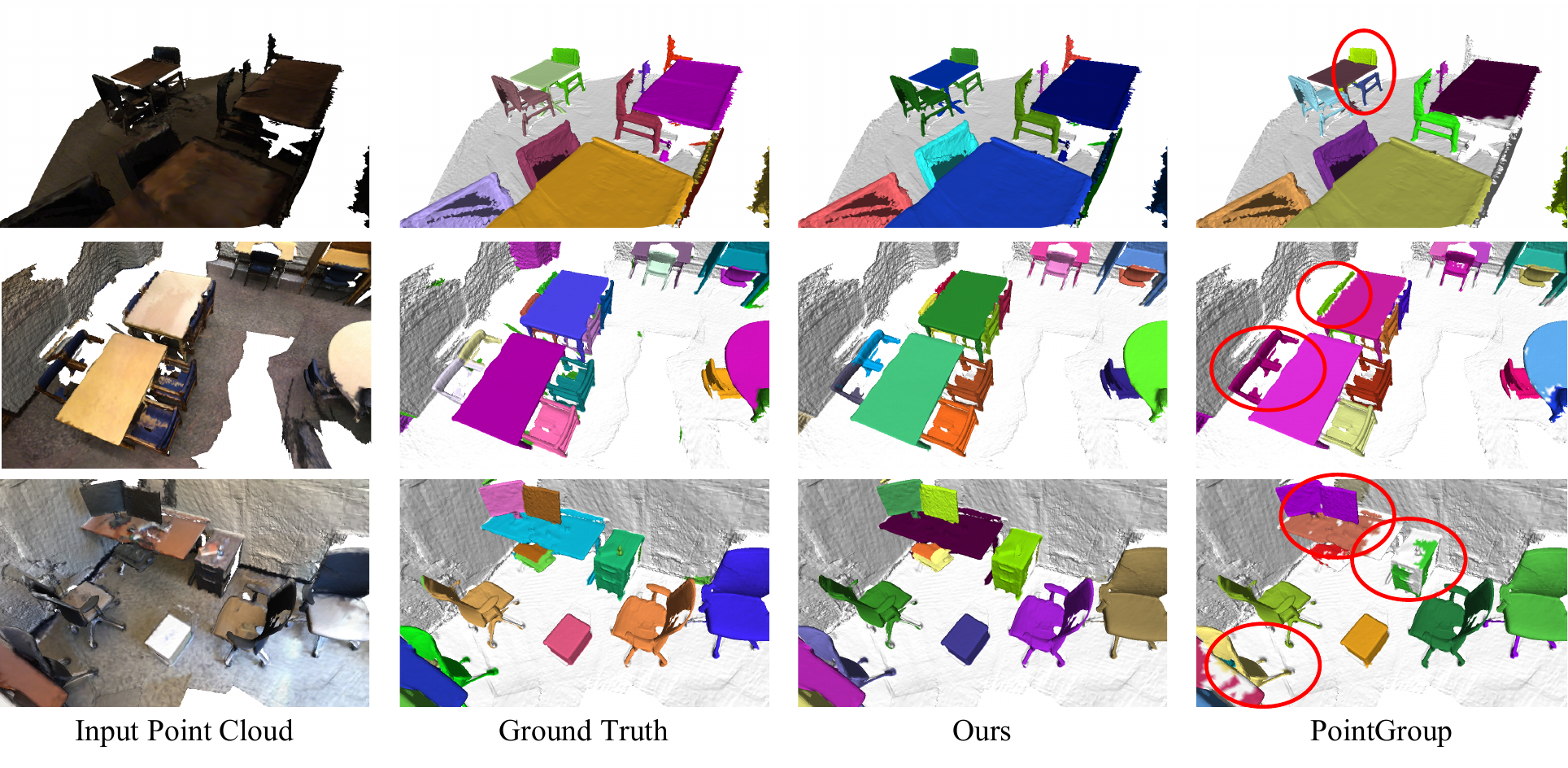}
	}
	
	\caption{Comparison of the results with PointGroup \cite{jiang2020pointgroup} (best viewed in color.). The ellipses highlight specific over-segmentation/joint errors. As we can see, \SexyName performs better than PointGroup.
	}
	\label{fig:compg}
\end{figure*}

\subsection{Datasets}
S3DIS contains 13 categories that are widespread in indoor scenes. The point cloud data is collected in 6 large-scale areas, covering more than 6000 $m^2$ with more than 215 million points. Following the evaluation protocols from previous methods \cite{wang2018sgpn, he2020eccvmemory}, 
we report the performance of both Area-5 and 6-fold sets. 
ScanNet \cite{dai2017scannet} is another large-scale benchmark for indoor scene analysis, which consists of 1613 scans with 40 categories in total. The dataset is split into 1201, 312, and 100 for training, evaluating, and testing, respectively. 
Following previous methods, we estimate the performance of instance segmentation on 18 common categories.
For the above two benchmarks, we apply a sparseconv-based backbone. 
To demonstrate the superiority of our proposed method, we extend the idea of dynamic convolution to a point-based method \cite{he2020eccvmemory} and evaluate the performance of instance segmentation on PartNet \cite{mo_2019_partnet}.
Different from the above two datasets, PartNet \cite{mo_2019_partnet} is a consistent
large-scale dataset with fine-grained and hierarchical part annotations. It consists of more than 570k part instances covering 24 object categories. Each category has three different levels of annotations. $Level$-1 has the coarsest and $Level$-3 has the finest annotations. 
Similar to \cite{he2020eccvembedding, yi2018gspn}, we select five categories that have the largest number of training examples for evaluation.

\subsection{Evaluation Metrics}
For ScanNetV2 and PartNet, we report the metric of mean average precision (mAP), which is widely used in the 2D image domain. AP@50 and AP@25 are also provided, having an IoU threshold of 0.5 and 0.25, respectively.
For S3DIS, we apply the metrics that are used in \cite{wang2019asis, he2020eccvembedding, he2020eccvmemory}: mConv, mWConv, mPrec, and mRec. 
mConv is defined as the mean instance-wise IoU. mWConv denotes the weighted version of mConv, where the weights are determined by the sizes of instances. mPrec and mRec denote the mean precision and recall with an IoU threshold of 0.5, respectively.

\subsection{Network Architectures.} 
\label{sec:network_arc}
The network architectures used in our experiments include \cite{graham2018sparseconv}, \cite{choy20194d}, and \cite{qi2017pointnetplusplus}. The first two are voxel-based and the last one is point-based. The network proposed by \cite{graham2018sparseconv} contains 7 blocks, each of which 
has a down-sampling factor of 2. The feature dimensions of the blocks are: 1$\times$c, 2$\times$c, $\cdots$, 7$\times$c, where $i$ controls the scaleability of the model. 
The default setting is identical with \cite{jiang2020pointgroup}, where $i$ is set to 16. We also report the performance of a larger model with $i$ set to 32, denoted as \SexyName-\text{L}. Minkowski-Engine\cite{choy20194d},is another framework for sparse convolution, which has been widely used in \cite{choy2019fully, choy2020high, gwak2020gsdn}. The performance of using Minkowski-Engine is denoted as \SexyName-\text{Mink}. For point-based architecture, we utilize PointNet++ \cite{qi2017pointnetplusplus}. \revise{Compared with voxel-based backbones, PointNet++ processes a limited number of points, due to the heavy computation on the neighboring searching. Thus, we conduct experiments on PartNet, a consistent large-scale dataset with fine-grained and hierarchical part annotations.}

\subsection{Ablation Studies}
In this section, we analyze the effect of each component in our proposed \SexyName. Performance is reported in terms of mAP, AP@50, and AP@25. All experiments are conducted with the same setting and training schedule, and are evaluated on ScanNetV2 \cite{dai2017scannet} validation set.

\revise{\subsubsection{Tradition CNN.}}
\revise{For the task of instance segmentation, two different clues are necessary: content and location. However, traditional CNNs/MLPs struggle at encoding this knowledge, as the weights are shared for all the instances and locations. Traditional CNNs/MLPs often follow the detect-then-segment pipeline for the task of instance segmentation. However, due to the unsatisfactory performance of indoor 3D detection, these methods often lead to sub-par performance and joint/fragmented instances. We use the detection results from VoteNet~\cite{qi2019deep}, and the method tries to classify foreground/background points. 
As shown in Table~\ref{tab:ablation_study}, this method obtains inferior performance due to the low accuracy in 3D detection.}

\subsubsection{Baseline.} 
We build our baseline by generating filters for every valid point without introducing any clustering operation and geometric embedding. One typical scan in ScanNetv2 usually contains 100K points, which is a heavy burden for both training and inference. We propose to randomly select 150 among foreground points for generating convolutional parameters, meaning outputting 150 instance proposals. The convolution operations are operated on all foreground points, without appending relative position information. As presented in Table \ref{tab:ablation_study}, the baseline model achieves 24.8, 43.8, and 56.4 in terms of mAP, AP@50, and AP@25, respectively. 

\begin{table}[!b]
	\small 
	\centering
	\addtolength{\tabcolsep}{-3.0pt}
	
	\begin{center}
		\renewcommand\arraystretch{1.25}
		\begin{tabular}{c|cccc|c|c|c}
			\toprule
			
			Method & Group  & PosEmb &CAD  &TF & mAP &AP@50 &AP@25  \\
			\toprule
			\revise{MLP/CNN} & & & & &\revise{16.7} &\revise{31.9} &\revise{50.6} \\
			Baseline & & & & &24.8 &43.8 &56.4 \\
			CondInst & & \checkmark& & &27.0 &44.7 &57.5 \\
			CondInst-cen & & \checkmark& & &28.1 &44.7 &57.1 \\
			&\checkmark & & & &29.4 &49.7 &66.3\\
			&\checkmark &\checkmark & & &31.8&52.9 &68.4\\
			&\checkmark &\checkmark &\checkmark & &34.1 &55.3 &69.5\\
			Ours&\checkmark &\checkmark &\checkmark &\checkmark &34.8 &55.7 &71.2\\
			\bottomrule
		\end{tabular}
	\end{center}
	\caption{Ablation studies on the components of our proposed method. We evaluate the performance on the ScanNetV2 \cite{dai2017scannet} validation set. $\textbf{Group}$ indicates that the dynamic filters are generated by gathering homogenous points that share the semantic labels and have close centroids votes. \textbf{PosEmb} refers to the position embeddings $f_\text{pos}$. $\textbf{CAD}$ denotes the category-aware decoding that the convolution in the decoding process is only operated on category-specific points, instead of all points. $\textbf{TF}$ refers to the light-weight transformer applied for the backbone. 
	}
	
	\label{tab:ablation_study}
\end{table}

\subsubsection{CondInst \cite{tian2020conditional} for 3D.} 
We implement CondInst \cite{tian2020conditional}, which has demonstrated its success in the 2D image domain. We implement CondInst in 3D point cloud by  adding position embeddings for each instance decoder.  Details can be found in Section \ref{section: instance_gen}.
With the same sampling strategy as used in baseline, CondInst boosts the mAP by 2.2\%, as presented in the second row in Table \ref{tab:ablation_study}, showing the significance of relative position in discriminating different instances.
Many methods for detection \cite{tian2019fcos, zhou2019objects} and instance segmentation \cite{tian2020conditional} in the 2D image domain constrain the predictions within the central areas of the instances by classifying these pixels as positive. 
However, it is hard for 3D point cloud, as points are sampled from the surface, leading to void space for central areas of instances. 
To demonstrate the difficulty of this selective prediction in the 3D domain, we define the positive points with a distance from centroids less than 0.2m and negative points with a distance larger than 0.6m. Other points are ignored during the training. The performance is shown in the third row in Table \ref{tab:ablation_study}, denoted as \text{CondInst-cen}. It achieves slightly better mAP (improve about 1.1\%), but is still far from satisfactory. 

\begin{figure}
	\centering
	{
		\includegraphics[width=7cm]{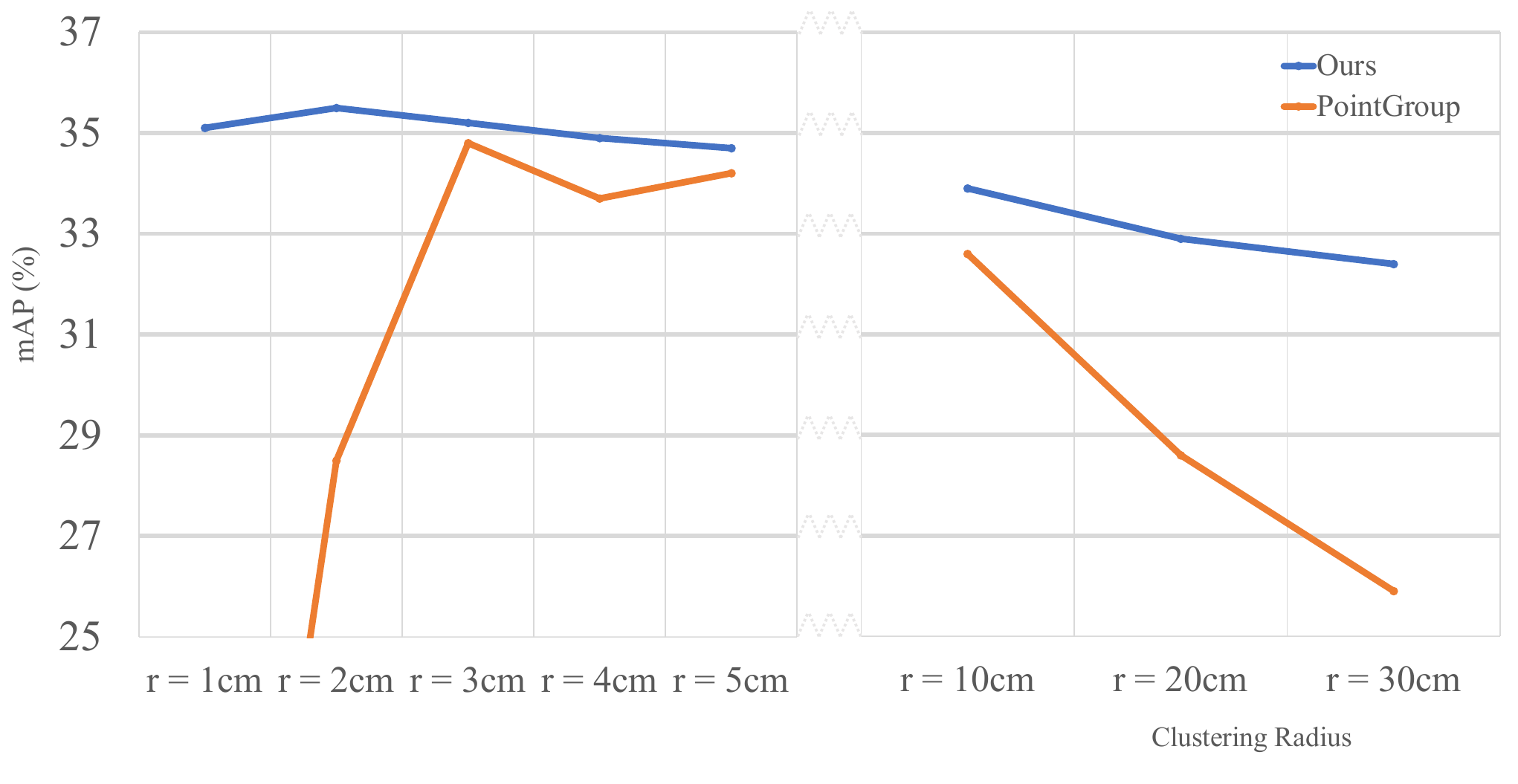}
	}
	
	\caption{\revise{The performance of instance segmentation with different clustering radius $r$. The performance of both common and extreme values is reported. All numbers for PointGroup are obtained from the paper or tested by the provided model. Unlike PointGroup, which is sensitive to the hyper-parameter and requires heuristic tuning, our method shows strong robustness.}
	}
	\label{fig:radius}
\end{figure}

\begin{figure}
	\centering
	{
		\includegraphics[width=7cm]{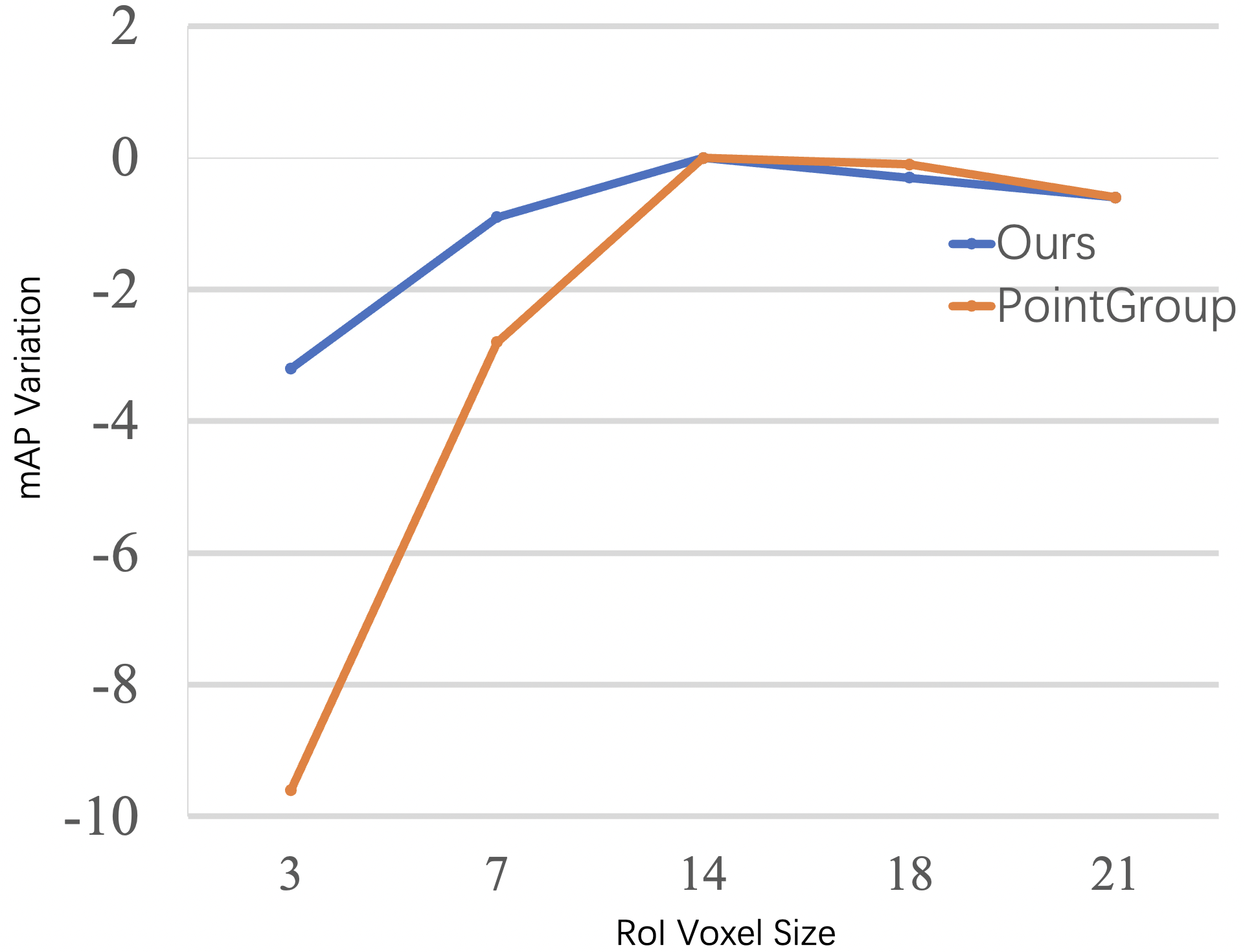}
	}
	
	\caption{\revise{The variation of the performance with different RoI voxel sizes. We select the model with a voxel size 14 as the baseline model.}
	}
	\label{fig:voxel_size}
\end{figure}

\subsubsection{Grouping Homogenous Points.} 
We propose to apply Algorithm \ref{alg: cluster} for grouping homogenous points that are defined in Sec. \ref{section: filter} to introduce a larger category-aware context. As presented in Table \ref{tab:ablation_study}, the model with a grouping step surpasses the baseline by a large margin in terms of all metrics. Besides, the grouping operation reduces the number of instance candidates (less than 100), lowering the optimization difficulties and releasing the heavy requirements for the hardware facilities. 
\revise{Different from PointGroup, which generates instance proposals aggregating homogenous points. The quality of the proposals is highly dependent on the performance of the two intermediate tasks: semantic segmentation and offset prediction. For example, the accuracy of the offset prediction is far from satisfactory, as illustrated in Figure 2. Consequently, the grouped proposals are likely to be over-segmented or under-segmented. In these cases, it is impossible to recover the accurate instance masks in the following steps, as only a scorenet is introduced. As a result, the quality of the proposal is sensitive to the performance of the two tasks. Instead of treating each cluster as a proposal, we utilize it to generate instance-related representations and decode the corresponding mask by simply applying several 1 $\times$1 convolution without complex heuristic settings.}

One of the most influential hyper-parameters in Algorithm \ref{alg: cluster}  is the the clustering radius. PointGroup \cite{jiang2020pointgroup} treats each cluster as an instance proposal, followed by a subnet for scoring. Thus the final performance is highly depending on the quality of the clusters. As illustrated in Figure \ref{fig:main}, minor changes of the radius lead to sub-par performance and joint/fragmented instances. We also provide quantitative results in Figure \ref{fig:radius}. Changing the values of radius $r$ from 3cm to 2cm drops mAP by 6.3\%, and 23.9\% by changing $r$ from 3cm to 1cm. The sensitivity of the value chosen makes it need to be carefully tuned for different datasets, demonstrating limited generalization capability to various scenes. In contrast, the performance of our proposed method \SexyName floats within 1\% under different settings, showing strong robustness to the searching radius $r$. \revise{To further demonstrate the superiority of the proposed method, we test the extreme cases by setting the searching radius to 10cm, 20cm, and 30cm. The results are present in Figure~\ref{fig:radius}. Our method can still maintain a reasonable fluctuation and outperforms PointGroup with different values chosen. For example, changing the radius from 3cm to 30cm, the mAP of our method drops by 2.8\%, while the mAP of PointGroup drops by about 9\%.}

\begin{figure}[htbp]
	\centering
	{
		\includegraphics[width=8cm]{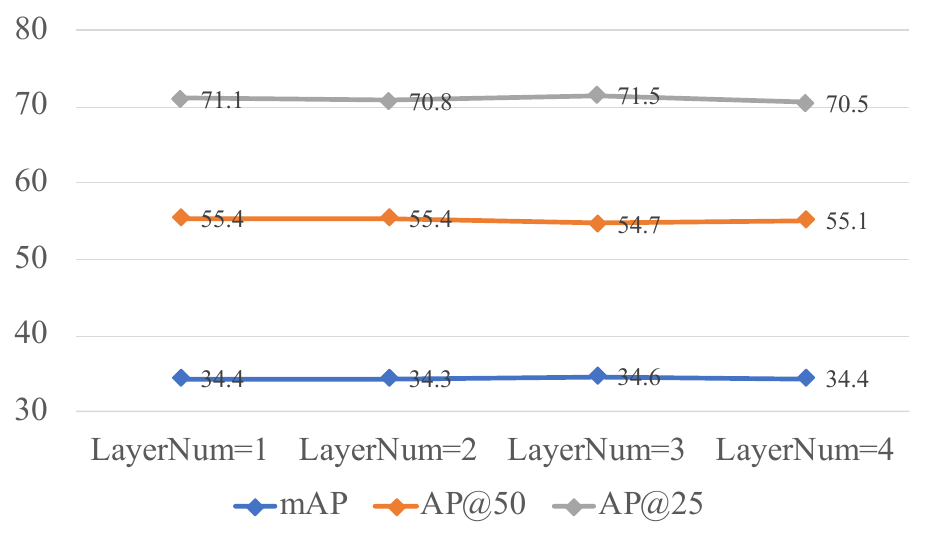}
	}
	\caption{The variation of the performance with different number of convolutional layers. 
	}
	\label{fig:layer_num}
\end{figure}

\begin{table}[]
	\small 
	\centering
	
	\begin{center}
		\begin{tabular}{c c c c}
			\toprule
			& mAP &mAP@50 &mAP@25\\
			\revise{Random} &\revise{38.7} &\revise{61.1} &\revise{72.9}\\
			\revise{Pre-training with \cite{PointContrast2020}} &\revise{39.3} &\revise{60.9} &\revise{73.7} \\
			\revise{Pre-training with \cite{hou2020exploring}} &\revise{39.1} &\revise{60.8} &\revise{73.8} \\

			\bottomrule
		\end{tabular}
	\end{center}
	
	\caption{\revise{The comparisons of the performance with different initialization strategies. The evaluation metrics of mAP, mAP@25, and mAP@50 are reported. Two popular pre-training strategies~\cite{PointContrast2020,hou2020exploring} are applied. The results are obtained by using a Minkowski-based backbone~\cite{choy20194d}. The dimension of the mask embedding is set to 16.}}
	
	\label{tab:init}
\end{table}

\subsubsection{Category-Aware Decoding.} 
In Algorithm \ref{alg: cluster}, center-shifted points that have identical category predictions are grouped. Thus, only specific category information is encoded in the weight generator. For decoding, we propose to apply convolution only for the points that have the same category with the filters. 
Instead of taking all instances into account, each filter only needs to discriminate one instance from one specific category, reducing the difficulties for learning. For example, if the filter is responsible for one specific chair, it only needs to segment out the chair from all other points that are classified as chair, rather than considering all unrelated categories, \eg desks, and sofa. 
As presented in Table\ref{tab:ablation_study}, adding category masks improves the mAP from 31.8\% to 34.1\%.

\subsubsection{Effectiveness of the Transformer.} 
As limited context and representation ability introduced by the sparse convolution, we propose to add a light-weighted transformer upon the bottleneck layer to capture the long-range dependencies and enhance the interactions among different points, while maintaining efficiency. The transformer module only contains the encoder part with a multi-head self-attention layer and a feed forward layer. 
As presented in Table \ref{tab:ablation_study}, the transformer brings about 0.7\% increase in terms of mAP.

\subsubsection{Can additional ScoreNet improve the performance?}
The method in \cite{jiang2020pointgroup} introduces an extra ScoreNet, which is a small sub-network to classify if the proposal is a valid instance. On the other hand, we select a naive approach by scoring the instance as the mean value of the mask scores among the foreground points. We conduct experiments by adding an extra ScoreNet and the results turn out only 0.3\% improvement, showing the discriminative capability of the instance-aware filters.


\subsubsection{Analysis on other hyper parameters.}

\begin{figure}[htbp]
	\centering
	{
		\includegraphics[width=7cm]{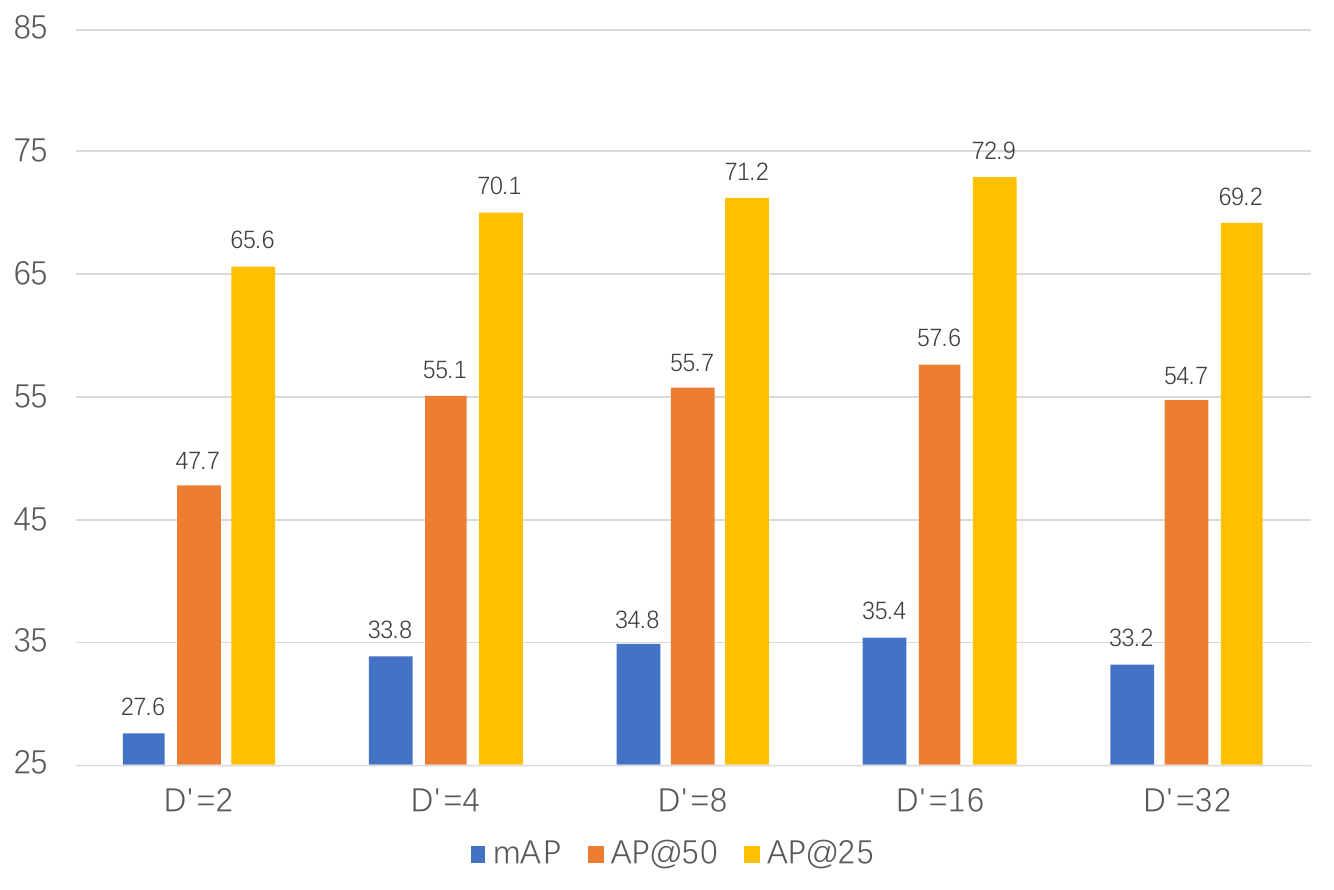}
	}
	\caption{The variation of the performance with different numbers of channels of the Mask Head.
	}
	\label{fig:dim}
\end{figure}

\textbf{The feature dimension of the mask head.} The feature $\textbf{F}_\text{mask}$ output by the mask head contains rich shape information and is shared by all instance decoders. We studied the significance of the dimension of $\textbf{F}_\text{mask}$. As show in Figure \ref{fig:dim}, we change $D'$ from 2 to 32. As $D'$ increases, the performance of all metrics gets better and will introduce more computational overheads. We set $D'$ to 16 in the following experiments, as it provides the best performance.

\textbf{The number of convolutional layers.} The output of the weight generator is a vector that contains all the convolutional parameters for the following convolution layers. In this part, we analyze the influence of the number of convolution layers. All results are trained for 30k steps, and the embedding dim $D'$ is set to 8. As can be seen from Figure \ref{fig:layer_num}, all evaluation metrics fluctuate within a reasonable scope (less than 1\%), showing that our method is robust to the number of the convolutional layers in the decoding. 

\revise{\textbf{The voxel size $g$ of the RoI pooling operation.} For each cluster $\mathcal{C}_\text{ctr}^z$, we voxelize it with a grid size of $g$. We evaluate the influence of the value chosen. As illustrated in Figure~\ref{fig:voxel_size}, the voxel size with a value of 14 obtains the best performance, when evaluated on the ScanNet dataset. We also report the variation with different RoI voxel sizes in terms of mAP. Compared with PointGroup~\cite{jiang2020pointgroup}, the accuracy of our proposed method fluctuates within a reasonable scope, demonstrating the robustness of our method to this heuristic setting.}


\begin{table}[!t]
	\small 
	\centering
	
	\begin{center}
		\begin{tabular}{r c c}
			\toprule
			
			\multicolumn{3}{c}{\textbf{3D Object Detection}}  \\ 
			\midrule
			ScanNetV2     & AP@25\% & AP@50\%  \\
			\midrule
			DSS~\cite{Song2016cvpr}      & 15.2  & 6.8  \\
			MRCNN 2D-3D~\cite{he2018maskrcnn}& 17.3  & 10.5 \\
			F-PointNet~\cite{qi2017frustum} & 19.8  & 10.8 \\
			GSPN~\cite{yi2018gspn}       & 30.6  & 17.7 \\
			3D-SIS~\cite{hou20193dsis}    & 40.2  & 22.5 \\
			VoteNet~\cite{qi2019deep}    & 58.6  & 33.5 \\
			PointGroup~\cite{jiang2020pointgroup} &56.8 &42.3 \\
			\textbf{\SexyName}                 &{58.9}  &{45.3} \\
			\revise{\textbf{\SexyName-L} }            &{\revise{62.0}}      &{\revise{47.9}} \\
			\textbf{\SexyName-Mink} &\textbf{63.2} &\textbf{49.3}\\ 
			
			\bottomrule
		\end{tabular}
	\end{center}
	
	\caption{3D object detection on the validation set of ScanNetV2~\cite{dai2017scannet}. We report per-class average precision (AP) with IoU thresholds of 25 \% and 50 \%. The performance of PointGroup \cite{jiang2020pointgroup} is evaluated with the provided model. We use the same backbone as \cite{jiang2020pointgroup} for a fair comparison. We also report the results by using Minkowski Engine\cite{choy20194d}, denoted as \SexyName-Mink.
	}
	
	\label{tab:object_detection}
\end{table}

\begin{table}[!t]
	
	\begin{center}
		\renewcommand\arraystretch{1.25}
		\begin{tabular}{r|c|c|c|c}
			\toprule
			\multicolumn{5}{c}{Test on Area 5} \\
			\hline
			Method  & mCov    & mWCov    &  mPrec & mRec \\
			
			\hline
			SGPN'18 ~\cite{wang2018sgpn}   &  32.7  & 35.5  &  36.0  & 28.7   \\
			
			ASIS'19 ~\cite{wang2019asis}   & 44.6  &  47.8 &  55.3 &  42.4 \\
			3D-BoNet'19 ~\cite{yang20193dbonet}   & -  &  - &  57.5 &  40.2 \\
			3D-MPA'20~\cite{Engelmann20CVPR}   & - &- &63.1 &58.0\\
			MPNet'20~\cite{he2020eccvmemory}      & {50.1} &{53.2} &{62.5} &{49.0} \\
			
			InsEmb'20~\cite{he2020eccvembedding}   &49.9 & 53.2 & 61.3 &48.5 \\
			PointGroup'20~\cite{jiang2020pointgroup}  &- &- & 61.9 & 62.1 \\
			\revise{\textbf{\SexyName-L}} &\textbf{63.5} &{64.6} &{64.3} &{64.2} \\
			\revise{\textbf{\SexyName-Mink}} &{63.2} & \textbf{64.7} &\textbf{64.8} &\textbf{64.6} \\

			\hline
			\hline
			\multicolumn{5}{c}{Test on 6-fold} \\
			\hline
			SGPN'18 ~\cite{wang2018sgpn} &  37.9  & 40.8  &  31.2  & 38.2   \\
			MT-PNet'19 ~\cite{phamjsis3dcvpr19}  &-   &-   &24.9 &- \\
			MV-CRF'19 ~\cite{phamjsis3dcvpr19} &- &- &36.3 &- \\
			ASIS'19 ~\cite{wang2019asis} &51.2 &55.1  & 63.6 & 47.5 \\
			3D-BoNet'19 ~\cite{yang20193dbonet} &- &- &65.6 &47.6 \\
			PartNet'19 ~\cite{mo_2019_partnet}  &- &- &56.4 &43.4 \\ 
			InsEmb'20\cite{he2020eccvembedding} &54.5 & 58.0 &67.2 &51.8\\
			MPNet'20 \cite{he2020eccvmemory}  &{55.8} &{59.7} &{68.4} &{53.7} \\
			PointGroup'20 \cite{jiang2020pointgroup} &- &- &69.6 &69.2 \\
			3D-MPA'20 \cite{Engelmann20CVPR} &- &- &66.7 & 64.1 \\
			\revise{\textbf{\SexyName-L}} &\textbf{72.8}  &{74.9}  &\textbf{71.7}  &{74.6} \\
			\revise{\textbf{\SexyName-Mink}} &{71.9} & \textbf{75.1} & {71.0} &\textbf{75.3} \\
			
			\bottomrule
		\end{tabular}
	\end{center}
	
	\caption{The results of instance segmentation on the S3DIS dataset. The performances on both Area-5 and 6-fold are reported. A comparison with previous top-performing approaches is presented. \revise{In DyCo3D-L, We use the same backbone as PointGroup.}
	}
	
	\label{tab:s3dis_ins_results}
\end{table}

\begin{table*}[!t]
	\resizebox{\textwidth}{!}{
		\renewcommand\arraystretch{1.25}
		\begin{tabular}{ r |cc|cccccccccccccccccc}
			\toprule
			&\textbf{AP@50} &\textbf{mAP}& \rotatebox{90}{cabinet} & \rotatebox{90}{bed} & \rotatebox{90}{chair} & \rotatebox{90}{sofa} & \rotatebox{90}{table} & \rotatebox{90}{door} & \rotatebox{90}{window} & \rotatebox{90}{bookshe.} & \rotatebox{90}{picture} & \rotatebox{90}{counter} & \rotatebox{90}{desk} & \rotatebox{90}{curtain} & \rotatebox{90}{fridge} & \rotatebox{90}{s.curtain} & \rotatebox{90}{toilet} & \rotatebox{90}{sink} & \rotatebox{90}{bath} & \rotatebox{90}{otherfu.} \\
			\midrule
			SegClu~\cite{hou20193dsis}&10.8 &- &10.4&11.9&15.5&12.8&12.4&10.1&10.1&10.3&0.0&11.7&10.4&11.4&0.0&13.9&17.2&11.5&14.2&10.5\\
			MRCNN~\cite{he2018maskrcnn}&9.1 &- &11.2&10.6&10.6&11.4&10.8&10.3&0.0&0.0&11.1&10.1&0.0&10.0&12.8&0.0&18.9&13.1&11.8&11.6\\
			SGPN~\cite{wang2018sgpn}&11.3 &- &10.1&16.4&20.2&20.7&14.7&11.1&11.1&0.0&0.0&10.0&10.3&12.8&0.0&0.0&48.7&16.5&0.0&0.0\\
			3D-SIS~\cite{hou20193dsis}&18.7&- &19.7&37.7&40.5&31.9&15.9&18.1&0.0&11.0&0.0&0.0&10.5&11.1&18.5&24.0&45.8&15.8&23.5&12.9\\
			MPNet~\cite{he2020eccvmemory}&31.0&- &-&-&-&-&-&-&-&-&-&-&-&-&-&-&-&-&-&-\\
			MTML~\cite{Jean2019mtml}&40.2&- &14.5&54.0&79.2&48.8&42.7&32.4&32.7&21.9&10.9&0.8&14.2&39.9&42.1&64.3&96.5&36.4&70.8&21.5\\
			3D-MPA~\cite{Engelmann20CVPR}&{59.1} &35.3 &{51.9}&{72.2}&{83.8}&{66.8}&{63.0}&{43.0}&{44.5}&{58.4}&{38.8}&{31.1}&{43.2}&
			{47.7}&\textbf{61.4}&\textbf{80.6}&\textbf{99.2}&{50.6}&\textbf{87.1}&{40.3}\\
			PointGroup~\cite{jiang2020pointgroup} &56.9 &34.8 &48.1 &69.6 &87.7 &71.5 &62.9 &42.0 &46.2 &54.9 & 37.7 &22.4 &41.6 & 44.9&37.2 &64.4 &98.3 &61.1 &80.5 &53.0 \\

			\textbf{\SexyName}&57.6 &35.4 &50.6 &{73.8} &84.4 & 72.1 &\textbf{69.9} &40.8 &44.5 &\textbf{62.4} & 34.8 &21.2 & 42.2 &37.0 & 41.6 & 62.7 &92.9 &61.6 & 82.6 &47.5 \\
			\hline

			\textbf{\SexyName-L} & {61.0} & \textbf{40.6} &{52.3} & 70.4 &{90.2} & 65.8 &69.6 & 40.5 & {47.2} &48.4 &\textbf{44.7} & {34.9} & \textbf{52.3} &47.5 &51.5 & 70.3 & 94.8 & \textbf{74.3} & 77.4 &{56.4} \\
			
			\textbf{\SexyName-Mink} &\textbf{61.1} &38.7 &\textbf{55.0} &\textbf{75.3} &\textbf{92.8} &\textbf{73.5} &{69.2} &\textbf{49.9} &\textbf{48.4} &{60.8} &33.4 & \textbf{36.9} &45.4 &\textbf{48.6} &48.4 &55.2 &98.3 & 66.3 & 80.5 &\textbf{61.2}\\

			\bottomrule
		\end{tabular}
	}
	\caption{Per class 3D instance segmentation on ScanNetV2~\cite{dai2017scannet} validation set. Both mAP and AP@50 are reported.}
	\label{tab:scannet_val}
\end{table*}

\subsubsection{Strategies of Initialization.}
The finding of fine-tuning models that pre-trained on a large well-annotated dataset (\eg ImageNet) can significantly boost the performance and accelerate the speed of convergence. Different from the 2D image domain, which contains a large scale of both fine- and coarse-level labels, there are still no suitable large-scale well-annotated datasets for pre-training on 3D point cloud. The emergence of self-supervised learning \cite{he_mocov1, chen2020mocov2, chen2020simple, chen2020big, wang2020DenseCL, Wuunsupervise} alleviates the situation. PointContrast\cite{PointContrast2020} is one of the pioneering works that explore self-supervised pre-training in 3D scenes. Hou \cite{hou2020exploring} improved the performance by optimizing the sampling strategy. In this section, we study the influence with different initialization: random and self-supervised pre-training, which are shown in Table \ref{tab:init}. All the results are obtained with $D'$ set to 8. The backbone is Minkovsiki-Engin \cite{choy20194d}, as pre-trained models are provided. With random initialization, our method achieves 34.9, 55.4, and 71.8, in terms of mAP, mAP@50, and mAP@25, respectively. The performance is boosted by 1.9 and 2.2 with approaches of \cite{PointContrast2020} and \cite{hou2020exploring}, respectively. 



\subsubsection{Analysis on the Efficiency.} 
\begin{figure}[htbp]
	\centering
	{
		\includegraphics[width=8cm]{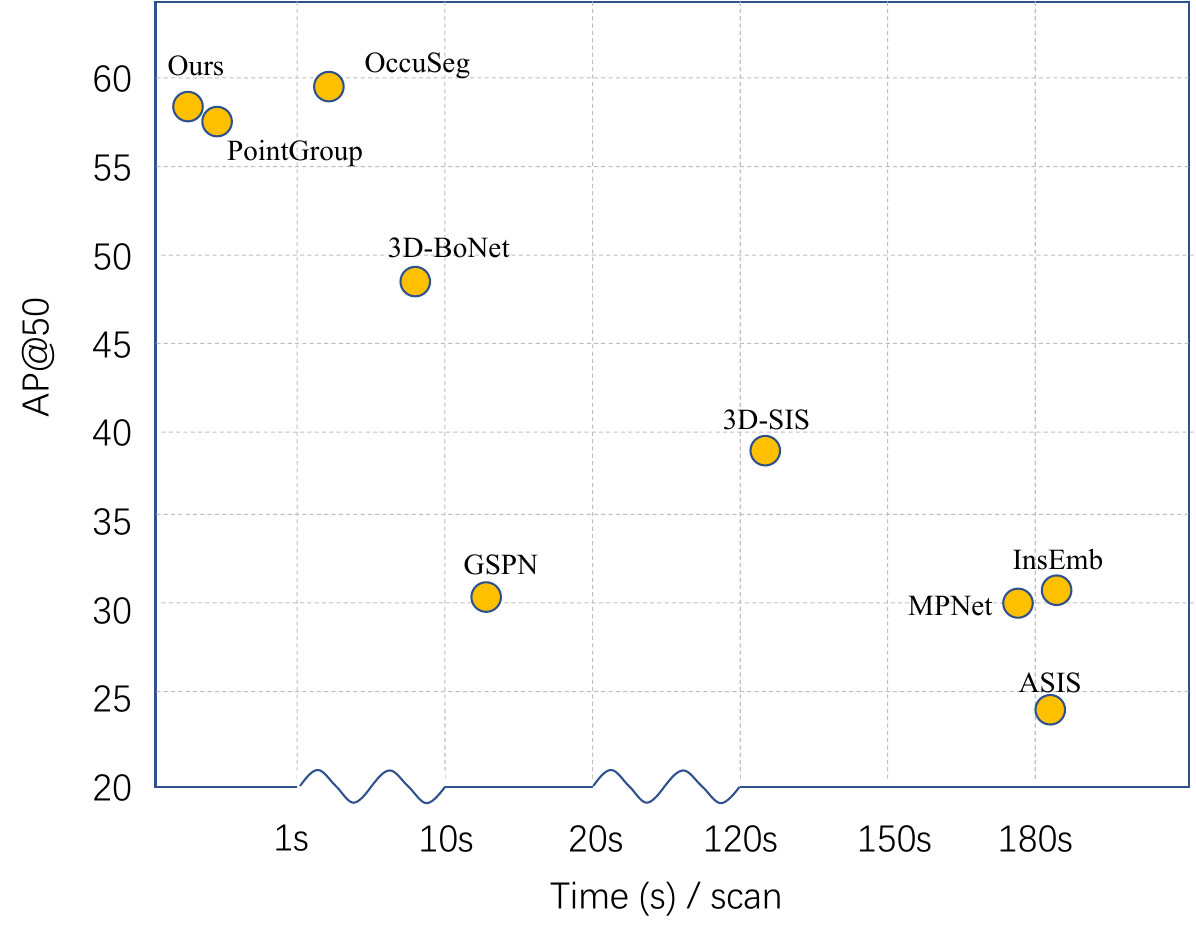}
	}
	\caption{\revise{The relation between inference time and AP@50. The results are obtained via testing on the ScanNet val set. Our proposed \SexyName achieves best performance while maintaining the highest efficiency.}
	}
	\label{fig:speed}
\end{figure}

Different from previous point-based approaches that require to split each scene as 1m $\times$ 1m blocks and apply a complex block merging algorithm \cite{wang2018sgpn, he2020eccvembedding, he2020eccvmemory, wang2019asis}, our method takes the whole scene as input. 
In addition, we also compare our \SexyName with PointGroup, which has shown its efficiency on large-scale scenes. We report the inference time that is averaged on the whole validation set. With the only post-processing step NMS, our method runs at 0.28s per scan on a 1080TI GPU, while the PointGroup runs at 0.39s with the same facility. Figure \ref{fig:speed} shows the trade-off between efficiency and effectiveness. Our proposed \SexyName achieves the best performance while maintaining a fast inference speed. 

\begin{table}[!]
	\centering
	
	\begin{center}
		\renewcommand\arraystretch{1.25}
		\begin{tabular}{c|ccc}
			\toprule
			
			\revise{PointNum} & \revise{DyCo3D}  & \revise{PointGroup} &\revise{OccuSeg} \\
			\revise{305916}       &          \revise{0.56s}     &     \revise{0.69s}        &    \revise{3.31s }      \\
			\revise{252726}        &           \revise{0.30s}    &   \revise{ 0.34s}       &   \revise{2.64s}        \\
			\revise{167915}        &          \revise{ 0.27s}     &    \revise{  0.38s}    &  \revise{2.10s}         \\
		%
			\bottomrule
		\end{tabular}
	\end{center}
	\caption{Comparison of the running time for different methods.}
	
	\label{tab:efficiency}
\end{table}

\subsection{Comparison with State-of-the-art Methods}
\textbf{Object Detection.} We report the performance of 3D object detection on the validation set of ScanNetV2. Following \cite{Engelmann20CVPR}, these boxes are obtained by fitting axis-aligned bounding boxes containing the instances.
As shown in Table \ref{tab:object_detection}, our method surpasses PointGroup\cite{jiang2020pointgroup} with the same backbone architecture by 3.1\% and 3.0\% in terms of AP@25 and AP@50, respectively, demonstrating the compactness of the instance masks. Qualitative results are shown in Figure \ref{tab:s3dis_ins_results}. Our proposed method outperforms the baseline model and can generate compact segmentation results. 
We also report the performance by applying Minkowski Engine \cite{choy20194d} with a ResNet16Unet, which achieves a much higher accuracy, denoted as \SexyName-Mink in Table \ref{tab:object_detection}.

\begin{figure}[htbp]
	\centering
	{
		\includegraphics[width=8.5cm]{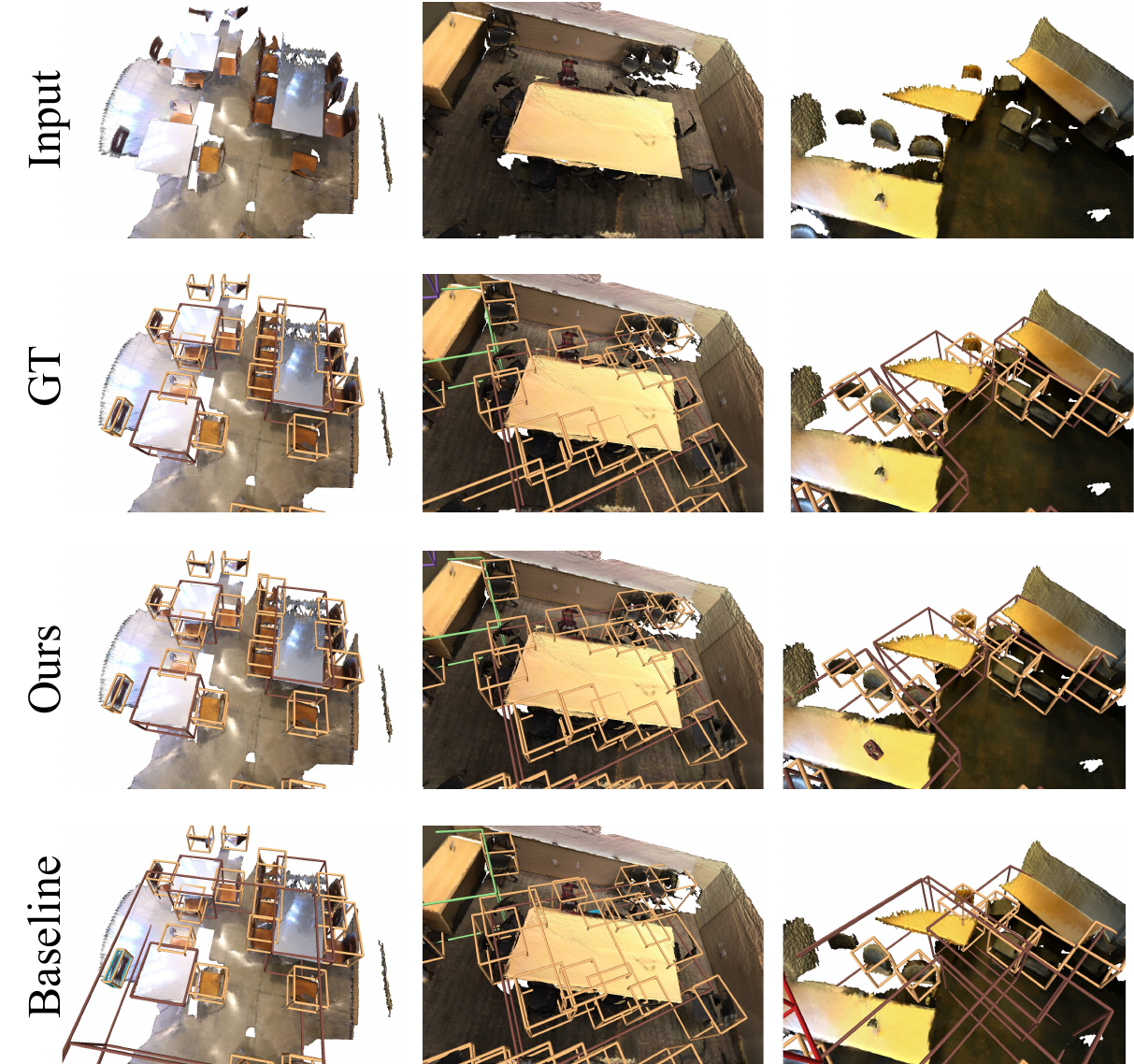}
	}
	\caption{The performance of 3D detection results, which are obtained by fitting axis-aligned bounding boxes containing the instances. The detection results demonstrate that our method can generate compact segmentation results. 
	}
	\label{fig:detection3d}
\end{figure}
\begin{figure*}[htbp]
	\centering
	{
		\includegraphics[width=16cm]{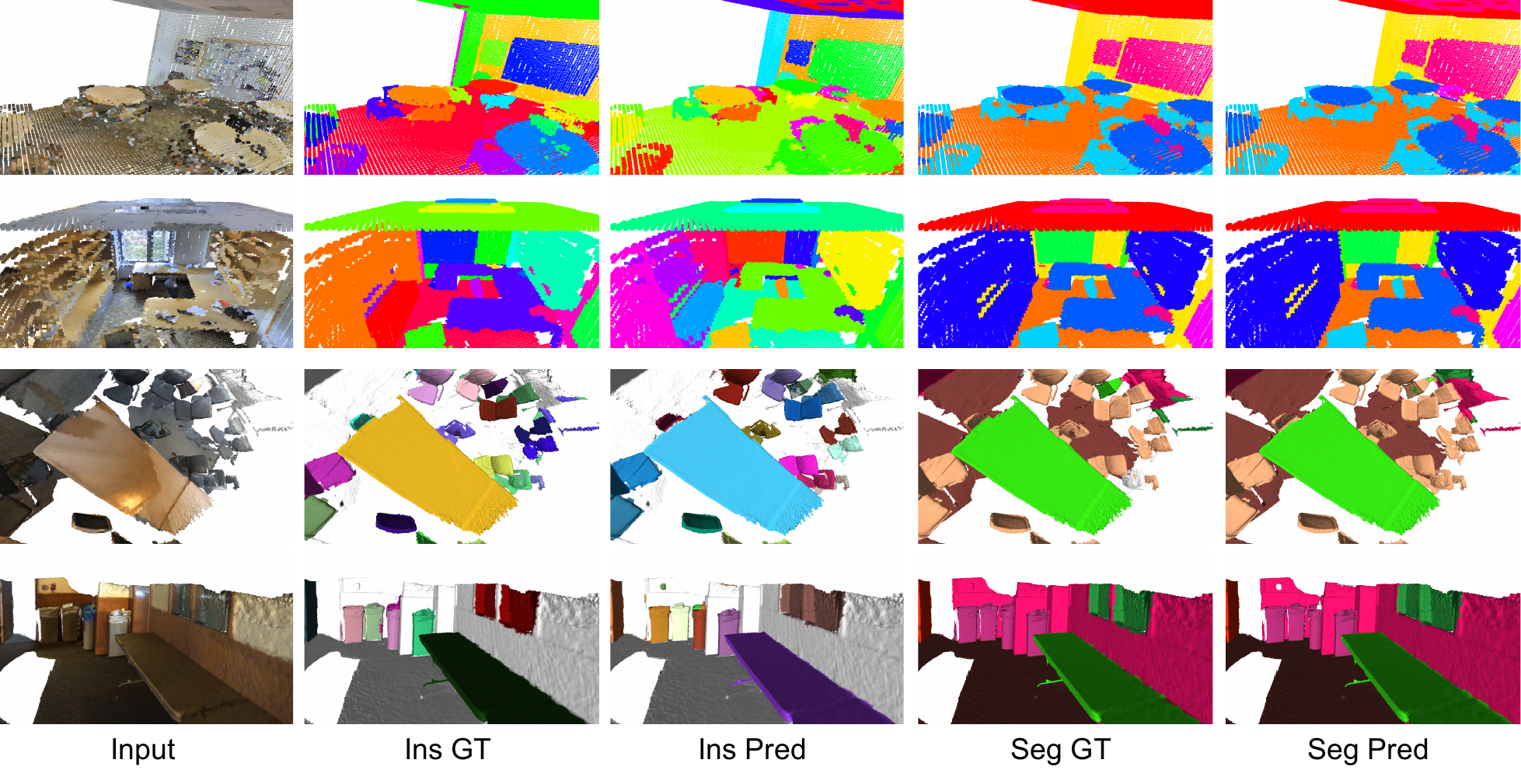}
	}
	
	\caption{Visualization of semantic and instance segmentation results on both S3DIS (top) and ScanNetv2 (bottom) benchmarks. Instances are presented with random colors.}
	\label{fig:results}
\end{figure*}
\begin{figure}[htbp]
	\centering
	{
		\includegraphics[width=8cm]{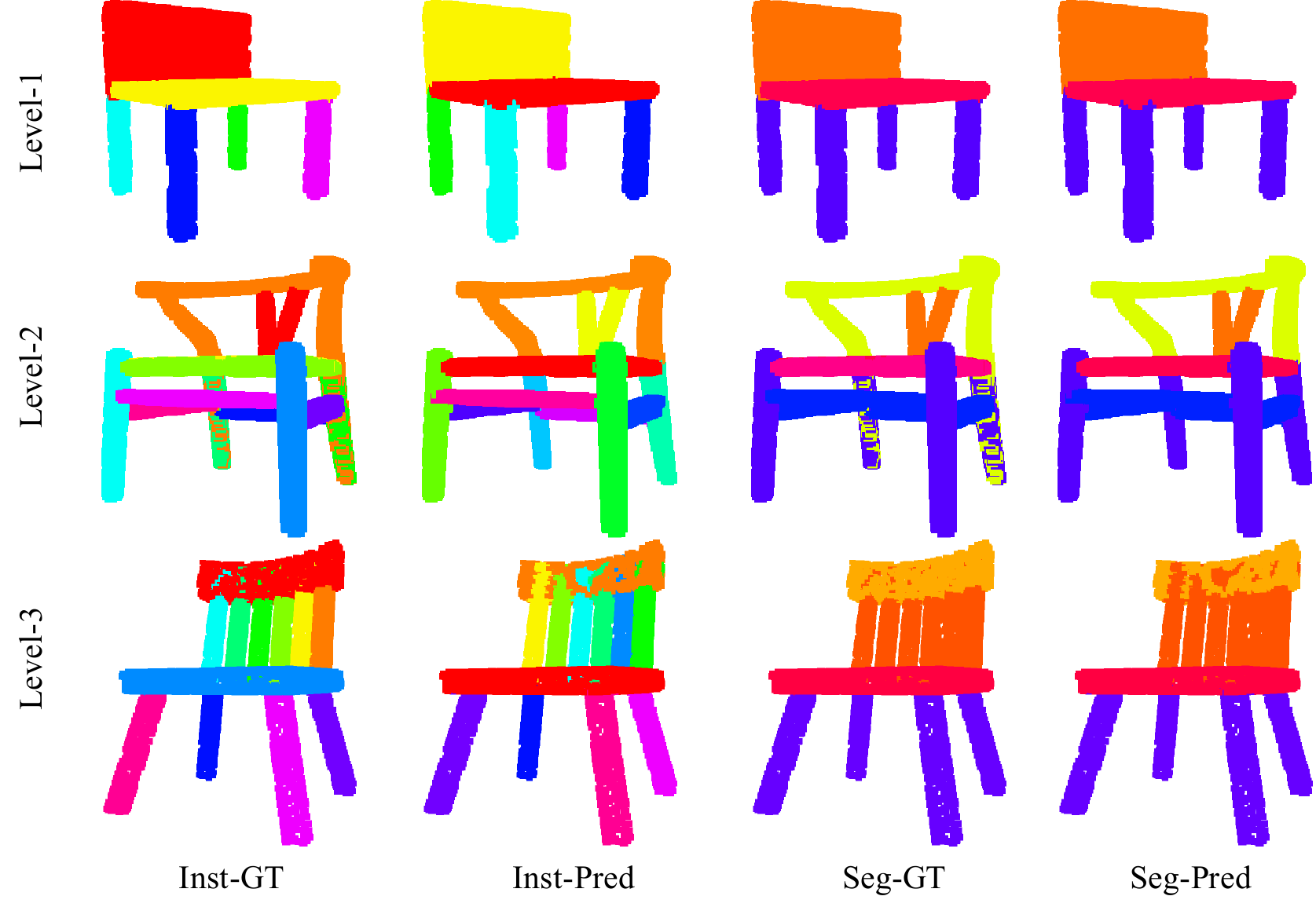}
	}
	\caption{The instance segmentation results on PartNet. Level-1 has the coarsest annotations, while level-3 has the finest annotations. Although there are some flaws in the label annotations (second row), our method can still provide accurate instance masks.
	}
	\label{fig:partnet}
\end{figure}
\textbf{Instance Segmentation on S3DIS.}
We report the performance of instance segmentation on the S3DIS benchmark, as shown in Table \ref{tab:s3dis_ins_results}. All these metrics are computed with the evaluation code provided by \cite{wang2019asis}. Results are tested on both Area-5 and 6-fold. 
Our method achieves the highest scores with all the evaluation metrics. When tested on Area-5, the results in terms of mPrec and mRec are 2.6\% and 2.1\% higher than PointGroup \cite{jiang2020pointgroup}. For 6-fold testing, \SexyName surpasses both 3D-MPA and PointGroup by a large margin, which is achieved under the same setting with ScanNetV2 without tuning heuristic hyper-parameters. Qualitative results are illustrated in Figure \ref{fig:results}.

\begin{table*}[!t]
	\centering
	\resizebox{\textwidth}{!}{
		\renewcommand\arraystretch{1.25}
		\begin{tabular}{ r |c|cccccccccccccccccc}
			
			\toprule[1pt]
			Method &  \textbf{AP@50} &\rotatebox{90}{bathtub}&\rotatebox{90}{bed}&\rotatebox{90}{bookshe.}&\rotatebox{90}{cabinet}&\rotatebox{90}{chair}&\rotatebox{90}{counter}&\rotatebox{90}{curtain}&\rotatebox{90}{desk}&\rotatebox{90}{door}&\rotatebox{90}{otherfu.}&\rotatebox{90}{picture}&\rotatebox{90}{refrige.}&\rotatebox{90}{s. curtain}&\rotatebox{90}{sink}&\rotatebox{90}{sofa}&\rotatebox{90}{table}&\rotatebox{90}{toilet}&\rotatebox{90}{window}\\
			\hline
			SGPN~\cite{wang2018sgpn} & 0.143 & 0.208 & 0.390 & 0.169 & 0.065 & 0.275 & 0.029 & 0.069 & 0.000 & 0.087 & 0.043 & 0.014 & 0.027 & 0.000 & 0.112 & 0.351 & 0.168 & 0.438 & 0.138 \\
			3D-BEVIS~\cite{elich20193dbevis} & 0.248 & 0.667 & 0.566 & 0.076 & 0.035 & 0.394 & 0.027 & 0.035 & 0.098 & 0.099 & 0.030 & 0.025 & 0.098 & 0.375 & 0.126 & 0.604 & 0.181 & 0.854 & 0.171 \\
			R-PointNet~\cite{yi2018gspn} &0.306 & 0.500 & 0.405 & 0.311 & 0.348 & 0.589 & 0.054 & 0.068 & 0.126 & 0.283 & 0.290 & 0.028 & 0.219 & 0.214 & 0.331 & 0.396 & 0.275 & 0.821 & 0.245 \\
			DPC~\cite{Engelmann20ICRA} & 0.355 & 0.500 & 0.517 & 0.467 & 0.228 & 0.422 & {0.133} & 0.405 & 0.111 & 0.205 & 0.241 & 0.075 & 0.233 & 0.306 & 0.445 & 0.439 & 0.457 & 0.974 & 0.23 \\
			3D-SIS~\cite{hou20193dsis} & 0.382 & 1.000 & 0.432 & 0.245 & 0.190 & 0.577 & 0.013 & 0.263 & 0.033 & 0.320 & 0.240 & 0.075 & 0.422 & 0.857 & 0.117 & 0.699 & 0.271 & 0.883 & 0.235 \\
			MASC~\cite{Liu2019masc} & 0.447 & 0.528 & 0.555 & 0.381 & 0.382 & 0.633 & 0.002 & 0.509 & 0.260 & 0.361 & 0.432 & 0.327 & 0.451 & 0.571 & 0.367 & 0.639 & 0.386 & 0.980 & 0.276 \\
			PanopticFusion~\cite{Narita2019iros} & 0.478 & 0.667 & 0.712 & 0.595 & 0.259 & 0.550 & 0.000 & 0.613 & 0.175 & 0.250 & 0.434 & 0.437 & 0.411 & 0.857 & 0.485 & 0.591 & 0.267 & 0.944 & 0.35 \\
			3D-BoNet~\cite{yang20193dbonet} & 0.488 & 1.000 & 0.672 & 0.590 & 0.301 & 0.484 & 0.098 & 0.620 & 0.306 & 0.341 & 0.259 & 0.125 & 0.434 & 0.796 & 0.402 & 0.499 & 	0.513 & 0.909 & 0.439 \\
			MTML~\cite{Jean2019mtml} & 0.549 & 1.000 & {0.807} & 0.588 & 0.327 & 0.647 & 0.004 & {0.815} & 0.180 & 0.418 & 0.364 & 0.182 & 0.445 & 1.000 & 0.442 & 0.688 & {0.571} & {1.000} & 0.396 \\
			PointGroup~\cite{jiang2020pointgroup} & {0.636} & {1.000} & 0.765 & {0.624} & {0.505} & {0.797} & 0.116 & 0.696 & {0.384} & {0.441} & {0.559} & {0.476} & {0.596} & {1.000} & {0.666} & {0.756} & 0.556 & 0.997 & {0.513} \\
			3D-MPA~\cite{Engelmann20CVPR} & {0.611} &1.000 &0.833  &0.765  &0.526  &0.756  &0.136  &0.588  &0.470  &0.438  &0.432  &0.358  &0.650  &0.857  &0.429  &0.765  &0.557  &1.000  &0.430  \\

			\revise{OccuSeg~\cite{han2020occuseg}$^*$} &\revise{0.634 (0.672)} &\revise{1.000} &\revise{0.758} &\revise{0.682} &\revise{0.576} &\revise{0.842} &\revise{0.477} &\revise{0.504} &\revise{0.524} &\revise{0.567} & \revise{0.585} & \revise{0.451} & \revise{0.577} & \revise{1.000} & \revise{0.571} & \revise{0.797} & \revise{0.563} & \revise{1.000} &\revise{0.467} \\
			
			\revise{PE~\cite{Zhang_2021_CVPR}} &\revise{0.645} &\revise{1.000} &\revise{0.773} &\revise{0.798} &\revise{0.538} &\revise{0.786} &\revise{0.088} &\revise{0.799} &\revise{0.350} &\revise{0.435} & \revise{0.547} & \revise{0.545} & \revise{0.646} & \revise{0.933} & \revise{0.562} & \revise{0.761} & \revise{0.556} & \revise{0.997} &\revise{0.501} \\
			
			\revise{\textbf{\SexyName-L}}&\revise{0.648}  &\revise{1.000} &\revise{0.759}  &\revise{0.554}  &\revise{0.595}  &\revise{0.819}  &\revise{0.309}  &\revise{0.562}  &\revise{0.460}  &\revise{0.505}  &\revise{0.555}  &\revise{0.446}  &\revise{0.593}  &\revise{1.000}  &\revise{0.558}  &\revise{0.787}  &\revise{0.552}  &\revise{1.000}  &\revise{0.605}  \\
			\midrule
		
			\revise{SSTNet~\cite{liang2021instance}} &\revise{0.698} &\revise{1.00} &\revise{0.697} &\revise{0.888}  &\revise{0.556} &\revise{0.803}  &\revise{0.387}  &\revise{0.626}  &\revise{0.417}  &\revise{0.556}  &\revise{0.585}  &\revise{0.702}  &\revise{0.600}  &\revise{1.000}  &\revise{0.824}  &\revise{0.720}  &\revise{0.692}  &\revise{1.000}  &\revise{0.509} \\
			\revise{HAIS~\cite{Chen_HAIS_2021_ICCV}} &\revise{0.699} &\revise{1.00} &\revise{0.849} &\revise{0.820}  &\revise{0.675} &\revise{0.808}  &\revise{0.279}  &\revise{0.757}  &\revise{0.465}  &\revise{0.517}  &\revise{0.596}  &\revise{0.559}  &\revise{0.600}  &\revise{1.000}  &\revise{0.654}  &\revise{0.767}  &\revise{0.676}  &\revise{0.994}  &\revise{0.560} \\

			\bottomrule[1pt]
		\end{tabular}
	}
	\caption{3D instance segmentation results on ScanNetV2 testing set with AP@50 scores on 18 categories. The result of OccuSeg~\cite{han2020occuseg} in brackets is from the benchmark website, while the other one is reported in the paper. \revise{The last two methods are from recently published papers. HAIS~\cite{Chen_HAIS_2021_ICCV} addressed the problem of over- and under-segmentation of PointGroup by introducing an extra filtering and scoring network. SSTNet~\cite{liang2021instance} introduced an complicated and time-consuming preprocessing step by building a semantic tree with superpoints.}
	}

	\label{tab:scannet_test}
\end{table*}

\textbf{Instance Segmentation results on ScanNetV2.}
We report the results of instance segmentation on both validation and testing sets of ScanNetV2, as presented in Table \ref{tab:scannet_val} and Table \ref{tab:scannet_test}, respectively.
On validation set, we report the metrics of AP@50 and mAP. The performance on two backbone architectures \cite{graham2018sparseconv, choy20194d} are evaluated. We also conduct experiments with a larger model capacity, denoted as \SexyName-L. Details of \SexyName-L can be found in Sec. \ref{sec:network_arc}. For a fair comparison, \SexyName-Mink surpasses 3D-MPA by 2.0\% and 3.4\% in terms of AP@50 and mAP, respectively. Our method also obtains better performance than \cite{jiang2020pointgroup} on both metrics. 
We also report the performance of \SexyName on the test set, as shown in Table \ref{tab:scannet_test}. 

\begin{table*}[htbp]
	
	\begin{center}
		\resizebox{\textwidth}{!}{
			\renewcommand\arraystretch{1.25}
			\begin{tabular}{r|ccccc|ccccc|ccccc|c}
				\toprule
				\multirow{2}{*}{Method} & \multicolumn{5}{c|}{Level1}            & \multicolumn{5}{c|}{Level 2}           & \multicolumn{5}{c}{Level 3}    &\multicolumn{1}{|c}{\revise{Ave}}        \\ \cline{2-17}
				& Chair & Storage & Table & Lamp  &\revise{Ave} & Chair & Storage & Table & Lamp &\revise{Ave}  & Chair & Storage & Table & Lamp &\revise{Ave}  &\\\hline 
				
				SGPN \cite{wang2018sgpn}                    & 72.4  & 32.9    & 49.2  & 32.7 &\revise{46.8}  & 25.4  & 30.5    & 18.9  & 21.7   &\revise{24.1}  & 19.4  & 21.5    & 14.6  & 14.4 &\revise{17.5} &\revise{29.5}  \\
				PartNet \cite{mo_2019_partnet}          & 74.4  & {45.2}    & 54.2  & {37.2} &\revise{52.8}  & 35.5  & 35.0    & 31.0  & 26.9  &\revise{29.6}  & 29.0  & 27.5    & 23.9  & 18.7 &\revise{24.8}  &\revise{36.5}\\
				GSPN \cite{yi2018gspn}                         & -     & -       & -     & - &-    & -    & -     & -     & -  &-  & 26.8  & 26.7    & 21.9  & 18.3 &\revise{23.4}   &-  \\
				ASIS \cite{yi2018gspn}                         & 77.1     & 43.2       & 55.0     & 34.1   & \revise{52.4}    & 36.0     & 35.5       & 31.3     & 24.8  &\revise{31.9}     & 28.9  & 27.4    & 22.8  & 19.1 &\revise{24.6} &\revise{36.3} \\
				InsEmb \cite{he2020eccvembedding}                   & 79.5 & 44.2    & \textbf{56.1}  & 36.1 & \revise{52.7} & {38.6}  & {37.1}    & \textbf{33.0}  & {26.9}  &\revise{33.9}  & {31.2}  & {28.9}    & {25.5}  & {19.4} &\revise{26.3} &\revise{38.1} \\
				\revise{ProbEmb~\cite{Zhang_2021_CVPR}} &\revise{77.1} &\textbf{\revise{47.3}} &\revise{40.3} &\revise{37.1} & \revise{48.0} &\revise{38.6} &\revise{42.0} &\revise{31.5} &\revise{31.0} &\revise{35.8} &\revise{34.7} &\textbf{\revise{34.2}} &\revise{25.5} &\revise{20.3} &\revise{28.7}  &\revise{38.3} \\
				
				Ours & \textbf{81.0}  & 44.5    &55.0   &\textbf{37.3} & \revise{54.5} & \textbf{41.3} &\textbf{38.9}   &{32.5}  &\textbf{28.8}  & \revise{35.4}   & \revise{\textbf{35.6}} & {30.4}  &\revise{\textbf{25.7}}   &\textbf{20.5} & \revise{28.1} &\revise{\textbf{39.3}}\\
				\bottomrule
			\end{tabular}
		}
	\end{center}
	\caption{Instance segmentation results on PartNet. We report part-category mAP (\%) under IoU threshold 0.5. There are three different levels for evaluation: coarse-grained level, middle-grained level and fine-grained level. We select five categories with the most data amount for training and evaluation. We put short lines for the levels that are not defined.}
	\label{tab:partnet}
\end{table*}

\textbf{Instance Segmentation on PartNet.}
The results of instance segmentation on PartNet are presented in Table \ref{tab:partnet}. Level-1 has the coarsest annotations and level-3 refers to the finest annotations. Similar to \cite{he2020eccvembedding}, the performance on five categories is reported. Different levels of each category are trained separately. Our method has achieved
state-of-the-art results on most categories and levels, improving the performance substantially. Results can be seen in Figure \ref{fig:partnet}. The consistent improvement on both volumetric and point-based backbones demonstrates the effectiveness of our proposed method.

	\section{Conclusion}
Achieving robustness to the inevitable variation in the data has been one of the ongoing challenges in the task of 3D point cloud segmentation.  We have shown here that dynamic convolution offers a mechanism by which to have the segmentation method actively respond to the characteristics of the data, and that this does in-fact improve robustness.  It also allows devising an approach that avoids many other pitfalls associated with bottom-up methods. The particular dynamic-convolution-based method that we have proposed, \SexyName, not only achieves state-of-the-art results, it offers improved efficiency over existing methods.



	%

	\balance
	\bibliographystyle{ieeetr}
	\bibliography{egbib}

\begin{thebibliography}{100}

\bibitem{jiang2020pointgroup}
L.~Jiang, H.~Zhao, S.~Shi, S.~Liu, C.-W. Fu, and J.~Jia, ``Pointgroup: Dual-set
  point grouping for 3d instance segmentation,'' in {\em IEEE Conf. Comput.
  Vis. Pattern Recog.}, 2020.

\bibitem{shi2019pointrcnn}
S.~Shi, X.~Wang, and H.~Li, ``Pointrcnn: 3d object proposal generation and
  detection from point cloud,'' in {\em IEEE Conf. Comput. Vis. Pattern
  Recog.}, 2019.

\bibitem{shi2020part}
S.~Shi, Z.~Wang, J.~Shi, X.~Wang, and H.~Li, ``From points to parts: 3d object
  detection from point cloud with part-aware and part-aggregation network,''
  {\em IEEE Trans. Pattern Anal. Mach. Intell.}, 2020.

\bibitem{shi2020pv}
S.~Shi, C.~Guo, L.~Jiang, Z.~Wang, J.~Shi, X.~Wang, and H.~Li, ``Pv-rcnn:
  Point-voxel feature set abstraction for 3d object detection,'' in {\em IEEE
  Conf. Comput. Vis. Pattern Recog.}, 2020.

\bibitem{yang2019std}
Z.~Yang, Y.~Sun, S.~Liu, X.~Shen, and J.~Jia, ``{STD}: Sparse-to-dense 3d
  object detector for point cloud,'' in {\em Int. Conf. Comput. Vis.}, 2019.

\bibitem{Shi_2020_CVPR}
W.~Shi and R.~Rajkumar, ``Point-gnn: Graph neural network for 3d object
  detection in a point cloud,'' in {\em IEEE Conf. Comput. Vis. Pattern
  Recog.}, 2020.

\bibitem{Li_cvpr_frodo}
K.~Li, M.~Rünz, M.~Tang, L.~Ma, C.~Kong, T.~Schmidt, I.~Reid, L.~Agapito,
  J.~Straub, S.~Lovegrove, and R.~Newcombe, ``Frodo: From detections to 3d
  objects,'' in {\em IEEE Conf. Comput. Vis. Pattern Recog.}, 2020.

\bibitem{lin2019photometric}
C.-H. Lin, O.~Wang, B.~C. Russell, E.~Shechtman, V.~G. Kim, M.~Fisher, and
  S.~Lucey, ``Photometric mesh optimization for video-aligned 3d object
  reconstruction,'' in {\em IEEE Conf. Comput. Vis. Pattern Recog.}, 2019.

\bibitem{sun_pix3d}
X.~Sun, J.~Wu, X.~Zhang, Z.~Zhang, C.~Zhang, T.~Xue, J.~B. Tenenbaum, and W.~T.
  Freeman, ``Pix3d: Dataset and methods for single-image 3d shape modeling,''
  in {\em IEEE Conf. Comput. Vis. Pattern Recog.}, 2018.

\bibitem{Mescheder2019CVPR}
L.~Mescheder, M.~Oechsle, M.~Niemeyer, S.~Nowozin, and A.~Geiger, ``Occupancy
  networks: Learning 3d reconstruction in function space,'' in {\em IEEE Conf.
  Comput. Vis. Pattern Recog.}, 2019.

\bibitem{Peng2020ECCV}
S.~Peng, M.~Niemeyer, L.~Mescheder, M.~Pollefeys, and A.~Geiger,
  ``Convolutional occupancy networks,'' in {\em Eur. Conf. Comput. Vis.}, 2020.

\bibitem{he2018maskrcnn}
K.~He, G.~Gkioxari, P.~Doll\'ar, and R.~Girshick, ``{Mask R-CNN},'' in {\em
  Int. Conf. Comput. Vis.}, 2017.

\bibitem{chen2020cascade}
K.~Chen, J.~Pang, J.~Wang, Y.~Xiong, X.~Li, S.~Sun, W.~Feng, Z.~Liu, J.~Shi,
  W.~Ouyang, and C.~C.~L. andDahua Lin, ``Hybrid task cascade for instance
  segmentation,'' in {\em IEEE Conf. Comput. Vis. Pattern Recog.}, 2019.

\bibitem{chen2020blendmask}
H.~Chen, K.~Sun, Z.~Tian, C.~Shen, Y.~Huang, and Y.~Yan, ``{BlendMask}:
  Top-down meets bottom-up for instance segmentation,'' in {\em IEEE Conf.
  Comput. Vis. Pattern Recog.}, 2020.

\bibitem{tian2020conditional}
Z.~Tian, C.~Shen, and H.~Chen, ``Conditional convolutions for instance
  segmentation,'' in {\em Eur. Conf. Comput. Vis.}, 2020.

\bibitem{hou20193dsis}
J.~Hou, A.~Dai, and M.~Nie{\ss}ner, ``{3D-SIS}: 3d semantic instance
  segmentation of rgb-d scans,'' in {\em IEEE Conf. Comput. Vis. Pattern
  Recog.}, 2019.

\bibitem{Zhang_2021_CVPR}
B.~Zhang and P.~Wonka, ``Point cloud instance segmentation using probabilistic
  embeddings,'' in {\em IEEE Conf. Comput. Vis. Pattern Recog.}, 2021.

\bibitem{he2020eccvembedding}
T.~He, Y.~Liu, C.~Shen, X.~Wang, and C.~Sun, ``Instance-aware embedding for
  point cloud instance segmentation,'' in {\em Eur. Conf. Comput. Vis.}, 2020.

\bibitem{he2020eccvmemory}
T.~He, D.~Gong, Z.~Tian, and C.~Shen, ``Learning and memorizing representative
  prototypes for 3d point cloud semantic and instance segmentation,'' in {\em
  Eur. Conf. Comput. Vis.}, 2020.

\bibitem{Engelmann20CVPR}
F.~Engelmann, M.~Bokeloh, A.~Fathi, B.~Leibe, and M.~Nie{\ss}ner, ``{3D-MPA}:
  Multi proposal aggregation for 3d semantic instance segmentation,'' in {\em
  IEEE Conf. Comput. Vis. Pattern Recog.}, 2020.

\bibitem{Jean2019mtml}
J.~Lahoud, B.~Ghanem, M.~Pollefeys, and M.~R. Oswald, ``3d instance
  segmentation via multi-task metric learning,'' in {\em Int. Conf. Comput.
  Vis.}, 2019.

\bibitem{wang2019asis}
X.~Wang, S.~Liu, X.~Shen, C.~Shen, and J.~Jia, ``Associatively segmenting
  instances and semantics in point clouds,'' in {\em IEEE Conf. Comput. Vis.
  Pattern Recog.}, 2019.

\bibitem{wang2018sgpn}
W.~Wang, R.~Yu, Q.~Huang, and U.~Neumann, ``{SGPN}: Similarity group proposal
  network for 3d point cloud instance segmentation,'' in {\em IEEE Conf.
  Comput. Vis. Pattern Recog.}, 2018.

\bibitem{Liu2019masc}
C.~Liu and Y.~Furukawa, ``{MASC}: Multi-scale affinity with sparse convolution
  for 3d instance segmentation,'' {\em arXiv preprint arXiv:1902.04478}, 2019.

\bibitem{han2020occuseg}
L.~Han, T.~Zheng, L.~Xu, and L.~Fang, ``Occuseg: Occupancy-aware 3d instance
  segmentation,'' in {\em IEEE Conf. Comput. Vis. Pattern Recog.}, 2020.

\bibitem{dai2017scannet}
A.~Dai, A.~X. Chang, M.~Savva, M.~Halber, T.~Funkhouser, and M.~Nie{\ss}ner,
  ``Scannet: Richly-annotated 3d reconstructions of indoor scenes,'' in {\em
  IEEE Conf. Comput. Vis. Pattern Recog.}, 2017.

\bibitem{armeni2016s3dis}
I.~Armeni, O.~Sener, A.~R. Zamir, H.~Jiang, I.~Brilakis, M.~Fischer, and
  S.~Savarese, ``3d semantic parsing of large-scale indoor spaces,'' in {\em
  IEEE Conf. Comput. Vis. Pattern Recog.}, 2016.

\bibitem{yang20193dbonet}
B.~Yang, J.~Wang, R.~Clark, Q.~Hu, S.~Wang, A.~Markham, and N.~Trigoni,
  ``Learning object bounding boxes for 3d instance segmentation on point
  clouds,'' in {\em Adv. Neural Inform. Process. Syst.}, 2019.

\bibitem{yi2018gspn}
L.~Yi, W.~Zhao, H.~Wang, M.~Sung, and L.~J. Guibas, ``{GSPN}: Generative shape
  proposal network for 3d instance segmentation in point cloud,'' in {\em IEEE
  Conf. Comput. Vis. Pattern Recog.}, 2018.

\bibitem{debrabandere16dynamic}
B.~De~Brabandere, X.~Jia, T.~Tuytelaars, and L.~Van~Gool, ``Dynamic filter
  networks,'' in {\em Adv. Neural Inform. Process. Syst.}, 2016.

\bibitem{tian2019fcos}
Z.~Tian, C.~Shen, H.~Chen, and T.~He, ``{FCOS}: Fully convolutional one-stage
  object detection,'' in {\em Int. Conf. Comput. Vis.}, 2019.

\bibitem{tian2021fcos}
Z.~Tian, C.~Shen, H.~Chen, and T.~He, ``{FCOS}: A simple and strong anchor-free
  object detector,'' {\em IEEE Trans. Pattern Anal. Mach. Intell.}, 2021.

\bibitem{qi2019deep}
C.~R. Qi, O.~Litany, K.~He, and L.~J. Guibas, ``Deep hough voting for 3d object
  detection in point clouds,'' in {\em Int. Conf. Comput. Vis.}, 2019.

\bibitem{graham2018sparseconv}
B.~Graham, M.~Engelcke, and L.~van~der Maaten, ``3d semantic segmentation with
  submanifold sparse convolutional networks,'' in {\em IEEE Conf. Comput. Vis.
  Pattern Recog.}, 2018.

\bibitem{choy20194d}
C.~Choy, J.~Gwak, and S.~Savarese, ``4d spatio-temporal convnets: Minkowski
  convolutional neural networks,'' in {\em IEEE Conf. Comput. Vis. Pattern
  Recog.}, 2019.

\bibitem{cheng2020panoptic}
B.~Cheng, M.~Collins, Y.~Zhu, T.~Liu, T.~Huang, H.~Adam, and L.-C. Chen,
  ``Panoptic-deeplab: A simple, strong, and fast baseline for bottom-up
  panoptic segmentation,'' in {\em IEEE Conf. Comput. Vis. Pattern Recog.},
  2020.

\bibitem{deeplabv3plus2018}
L.-C. Chen, Y.~Zhu, G.~Papandreou, F.~Schroff, and H.~Adam, ``Encoder-decoder
  with atrous separable convolution for semantic image segmentation,'' in {\em
  Eur. Conf. Comput. Vis.}, 2018.

\bibitem{ross_2014_cvpr_rich}
R.~{Girshick}, J.~{Donahue}, T.~{Darrell}, and J.~{Malik}, ``Rich feature
  hierarchies for accurate object detection and semantic segmentation,'' in
  {\em IEEE Conf. Comput. Vis. Pattern Recog.}, 2014.

\bibitem{wang_2015_torward}
{Peng Wang}, {Xiaohui Shen}, {Zhe Lin}, S.~{Cohen}, B.~{Price}, and
  A.~{Yuille}, ``Towards unified depth and semantic prediction from a single
  image,'' in {\em IEEE Conf. Comput. Vis. Pattern Recog.}, 2015.

\bibitem{dai_2017_deformable}
J.~{Dai}, H.~{Qi}, Y.~{Xiong}, Y.~{Li}, G.~{Zhang}, H.~{Hu}, and Y.~{Wei},
  ``Deformable convolutional networks,'' in {\em Int. Conf. Comput. Vis.},
  2017.

\bibitem{Vaswani2017attention}
A.~Vaswani, N.~Shazeer, N.~Parmar, J.~Uszkoreit, L.~Jones, A.~N. Gomez,
  L.~Kaiser, and I.~Polosukhin, ``Attention is all you need,'' in {\em Adv.
  Neural Inform. Process. Syst.}, 2017.

\bibitem{Simonyan15}
K.~Simonyan and A.~Zisserman, ``Very deep convolutional networks for
  large-scale image recognition,'' in {\em Int. Conf. Learn. Represent.}, 2015.

\bibitem{He2016resnet}
K.~He, X.~Zhang, S.~Ren, and J.~Sun, ``Deep residual learning for image
  recognition,'' in {\em IEEE Conf. Comput. Vis. Pattern Recog.}, 2016.

\bibitem{dosovitskiy2020image}
A.~Dosovitskiy, L.~Beyer, A.~Kolesnikov, D.~Weissenborn, X.~Zhai,
  T.~Unterthiner, M.~Dehghani, M.~Minderer, G.~Heigold, S.~Gelly, J.~Uszkoreit,
  and N.~Houlsby, ``An image is worth 16x16 words: Transformers for image
  recognition at scale,'' in {\em Int. Conf. Learn. Represent.}, 2020.

\bibitem{qi2017pointnet}
C.~R. Qi, H.~Su, K.~Mo, and L.~J. Guibas, ``Pointnet: Deep learning on point
  sets for 3d classification and segmentation,'' in {\em IEEE Conf. Comput.
  Vis. Pattern Recog.}, 2017.

\bibitem{qi2017pointnetplusplus}
C.~R. Qi, L.~Yi, H.~Su, and L.~J. Guibas, ``Pointnet++: Deep hierarchical
  feature learning on point sets in a metric space,'' in {\em Adv. Neural
  Inform. Process. Syst.}, 2017.

\bibitem{Zhao2021_pointtransformer}
H.~Zhao, L.~Jiang, J.~Jia, P.~Torr, and V.~Koltun, ``Point transformer,'' {\em
  arXiv preprint arXiv:2012.09164}, 2020.

\bibitem{hu2020randla}
Q.~Hu, B.~Yang, L.~Xie, S.~Rosa, Y.~Guo, Z.~Wang, N.~Trigoni, and A.~Markham,
  ``Randla-net: Efficient semantic segmentation of large-scale point clouds,''
  in {\em IEEE Conf. Comput. Vis. Pattern Recog.}, 2020.

\bibitem{Wu_2019_CVPR}
W.~Wu, Z.~Qi, and L.~Fuxin, ``Pointconv: Deep convolutional networks on 3d
  point clouds,'' in {\em IEEE Conf. Comput. Vis. Pattern Recog.}, 2019.

\bibitem{xu2018spidercnn}
Y.~Xu, T.~Fan, M.~Xu, L.~Zeng, and Y.~Qiao, ``Spidercnn: Deep learning on point
  sets with parameterized convolutional filters,'' in {\em Eur. Conf. Comput.
  Vis.}, 2018.

\bibitem{thomas2019KPConv}
H.~Thomas, C.~R. Qi, J.-E. Deschaud, B.~Marcotegui, F.~Goulette, and L.~J.
  Guibas, ``Kpconv: Flexible and deformable convolution for point clouds,'' in
  {\em Int. Conf. Comput. Vis.}, 2019.

\bibitem{maturana2015voxnet}
D.~Maturana and S.~Scherer, ``Voxnet: A 3d convolutional neural network for
  real-time object recognition,'' in {\em Proc. IEEE Int. Conf. Intelligent
  Robots Syst.}, 2015.

\bibitem{wu20153dshapenet}
Z.~Wu, S.~Song, A.~Khosla, F.~Yu, L.~Zhang, X.~Tang, and J.~Xiao, ``{3D
  ShapeNets}: A deep representation for volumetric shapes,'' in {\em IEEE Conf.
  Comput. Vis. Pattern Recog.}, 2015.

\bibitem{Riegler2017OctNet}
G.~Riegler, A.~O. Ulusoy, and A.~Geiger, ``Octnet: Learning deep 3d
  representations at high resolutions,'' in {\em IEEE Conf. Comput. Vis.
  Pattern Recog.}, 2017.

\bibitem{song2016ssc}
S.~Song, F.~Yu, A.~Zeng, A.~X. Chang, M.~Savva, and T.~Funkhouser, ``Semantic
  scene completion from a single depth image,'' in {\em IEEE Conf. Comput. Vis.
  Pattern Recog.}, 2017.

\bibitem{su2015multiview}
H.~Su, S.~Maji, E.~Kalogerakis, and E.~Learned-Miller, ``Multi-view
  convolutional neural networks for 3d shape recognition,'' in {\em Int. Conf.
  Comput. Vis.}, 2015.

\bibitem{qi2016multiview}
C.~R. Qi, H.~Su, M.~Nie{\ss}ner, A.~Dai, M.~Yan, and L.~J. Guibas, ``Volumetric
  and multi-view cnns for object classification on 3d data,'' in {\em IEEE
  Conf. Comput. Vis. Pattern Recog.}, 2016.

\bibitem{dai20183dmv}
A.~Dai and M.~Nie{\ss}ner, ``{3DMV}: Joint 3d-multi-view prediction for 3d
  semantic scene segmentation,'' in {\em Eur. Conf. Comput. Vis.}, 2018.

\bibitem{chen_2017_multiview}
X.~{Chen}, H.~{Ma}, J.~{Wan}, B.~{Li}, and T.~{Xia}, ``Multi-view 3d object
  detection network for autonomous driving,'' in {\em IEEE Conf. Comput. Vis.
  Pattern Recog.}, 2017.

\bibitem{kanezaki2018_rotationnet}
A.~Kanezaki, Y.~Matsushita, and Y.~Nishida, ``Rotationnet: Joint object
  categorization and pose estimation using multiviews from unsupervised
  viewpoints,'' in {\em IEEE Conf. Comput. Vis. Pattern Recog.}, 2018.

\bibitem{Lang_2019_CVPR}
A.~H. Lang, S.~Vora, H.~Caesar, L.~Zhou, J.~Yang, and O.~Beijbom,
  ``Pointpillars: Fast encoders for object detection from point clouds,'' in
  {\em IEEE Conf. Comput. Vis. Pattern Recog.}, 2019.

\bibitem{Tat2018}
M.~Tatarchenko, J.~Park, V.~Koltun, and Q.-Y. Zhou., ``Tangent convolutions for
  dense prediction in {3D},'' in {\em IEEE Conf. Comput. Vis. Pattern Recog.},
  2018.

\bibitem{lang2019pointpillar}
A.~H. Lang, S.~Vora, H.~Caesar, L.~Zhou, J.~Yang, and O.~Beijbom,
  ``Pointpillars: Fast encoders for object detection from point clouds,'' in
  {\em IEEE Conf. Comput. Vis. Pattern Recog.}, 2019.

\bibitem{liu2019pvcnn}
Z.~Liu, H.~Tang, Y.~Lin, and S.~Han, ``Point-voxel cnn for efficient 3d deep
  learning,'' in {\em Adv. Neural Inform. Process. Syst.}, 2019.

\bibitem{shi2021pv}
S.~Shi, L.~Jiang, J.~Deng, Z.~Wang, C.~Guo, J.~Shi, X.~Wang, and H.~Li,
  ``Pv-rcnn++: Point-voxel feature set abstraction with local vector
  representation for 3d object detection,'' {\em arXiv preprint
  arXiv:2102.00463}, 2021.

\bibitem{qi2020imvotenet}
C.~R. Qi, X.~Chen, O.~Litany, and L.~J. Guibas, ``Imvotenet: Boosting 3d object
  detection in point clouds with image votes,'' in {\em IEEE Conf. Comput. Vis.
  Pattern Recog.}, 2020.

\bibitem{Yang_2020_CVPR}
Z.~Yang, Y.~Sun, S.~Liu, and J.~Jia, ``3dssd: Point-based 3d single stage
  object detector,'' in {\em IEEE Conf. Comput. Vis. Pattern Recog.}, 2020.

\bibitem{tang2020searching}
H.~Tang, Z.~Liu, S.~Zhao, Y.~Lin, J.~Lin, H.~Wang, and S.~Han, ``Searching
  efficient 3d architectures with sparse point-voxel convolution,'' in {\em
  Eur. Conf. Comput. Vis.}, 2020.

\bibitem{Lei_2019_octree}
H.~{Lei}, N.~{Akhtar}, and A.~{Mian}, ``Octree guided cnn with spherical
  kernels for 3d point clouds,'' in {\em IEEE Conf. Comput. Vis. Pattern
  Recog.}, 2019.

\bibitem{matheus_2018_eccv_tree}
M.~Gadelha, R.~Wang, and S.~Maji, ``Multiresolution tree networks for 3d point
  cloud processing,'' in {\em Eur. Conf. Comput. Vis.}, 2018.

\bibitem{Shu_2019_ICCV}
D.~W. Shu, S.~W. Park, and J.~Kwon, ``3d point cloud generative adversarial
  network based on tree structured graph convolutions,'' in {\em Int. Conf.
  Comput. Vis.}, 2019.

\bibitem{hu2021bpnet}
W.~Hu, H.~Zhao, L.~Jiang, J.~Jia, and T.-T. Wong, ``Bidirectional projection
  network for cross dimension scene understanding,'' in {\em IEEE Conf. Comput.
  Vis. Pattern Recog.}, 2021.

\bibitem{Gaddetpami}
R.~Gadde, V.~Jampani, R.~Marlet, and P.~V. Gehler, ``Efficient 2d and 3d facade
  segmentation using auto-context,'' {\em IEEE Trans. Pattern Anal. Mach.
  Intell.}, 2018.

\bibitem{Liangcontfuse}
M.~Liang, B.~Yang, S.~Wang, and R.~Urtasun, ``Deep continuous fusion for
  multi-sensor 3d object detection,'' in {\em Eur. Conf. Comput. Vis.}, 2018.

\bibitem{su18splatnet}
H.~Su, V.~Jampani, D.~Sun, S.~Maji, E.~Kalogerakis, M.-H. Yang, and J.~Kautz,
  ``{SPLATN}et: Sparse lattice networks for point cloud processing,'' in {\em
  IEEE Conf. Comput. Vis. Pattern Recog.}, 2018.

\bibitem{phamjsis3dcvpr19}
Q.-H. Pham, D.~T. Nguyen, B.-S. Hua, G.~Roig, and S.-K. Yeung, ``{JSIS3D}:
  Joint semantic-instance segmentation of 3d point clouds with multi-task
  pointwise networks and multi-value conditional random fields,'' in {\em IEEE
  Conf. Comput. Vis. Pattern Recog.}, 2019.

\bibitem{bra2017cvprdis}
B.~D. Brabandere, D.~Neven, and L.~V. Gool, ``Semantic instance segmentation
  with a discriminative loss function,'' in {\em IEEE Conf. Comput. Vis.
  Pattern Recog.}, 2017.

\bibitem{Pedro2014_ijcv}
P.~F. Felzenszwalb and D.~P. Huttenlocher, ``Efficient graph-based image
  segmentation,'' {\em Int. J. Comput. Vis.}, 2004.

\bibitem{vu2022softgroup}
T.~Vu, K.~Kim, T.~M. Luu, X.~T. Nguyen, and C.~D. Yoo, ``Softgroup for 3d
  instance segmentation on 3d point clouds,'' in {\em IEEE Conf. Comput. Vis.
  Pattern Recog.}, 2022.

\bibitem{liang2021instance}
Z.~Liang, Z.~Li, S.~Xu, M.~Tan, and K.~Jia, ``Instance segmentation in 3d
  scenes using semantic superpoint tree networks,'' in {\em IEEE Conf. Comput.
  Vis. Pattern Recog.}, 2021.

\bibitem{Chen_HAIS_2021_ICCV}
S.~Chen, J.~Fang, Q.~Zhang, W.~Liu, and X.~Wang, ``Hierarchical aggregation for
  3d instance segmentation,'' in {\em Int. Conf. Comput. Vis.}, 2021.

\bibitem{Chen_2020_CVPR_dynamic}
Y.~Chen, X.~Dai, M.~Liu, D.~Chen, L.~Yuan, and Z.~Liu, ``Dynamic convolution:
  Attention over convolution kernels,'' in {\em IEEE Conf. Comput. Vis. Pattern
  Recog.}, 2020.

\bibitem{wu2018pay}
F.~Wu, A.~Fan, A.~Baevski, Y.~Dauphin, and M.~Auli, ``Pay less attention with
  lightweight and dynamic convolutions,'' in {\em Int. Conf. Learn.
  Represent.}, 2019.

\bibitem{Konstantin_adaptis2019}
A.~K. Konstantin~Sofiiuk, Olga~Barinova, ``Adaptis: Adaptive instance selection
  network,'' in {\em Int. Conf. Comput. Vis.}, 2019.

\bibitem{zhou2019objects}
X.~Zhou, D.~Wang, and P.~Kr{\"a}henb{\"u}hl, ``Objects as points,'' {\em arXiv
  preprint arXiv:1904.07850}, 2019.

\bibitem{dosovitskiy2020}
A.~Dosovitskiy, L.~Beyer, A.~Kolesnikov, D.~Weissenborn, X.~Zhai,
  T.~Unterthiner, M.~Dehghani, M.~Minderer, G.~Heigold, S.~Gelly, J.~Uszkoreit,
  and N.~Houlsby, ``An image is worth 16x16 words: Transformers for image
  recognition at scale,'' {\em arXiv preprint arXiv:2010.11929}, 2020.

\bibitem{Nicolas2020detr}
N.~Carion, F.~Massa, G.~Synnaeve, A.~K. Nicolas~Usunier, and S.~Zagoruyko,
  ``End-to-end object detection with transformers,'' {\em arXiv preprint
  arXiv:2010.11929}, 2020.

\bibitem{mo_2019_partnet}
K.~Mo, S.~Zhu, A.~X. Chang, L.~Yi, S.~Tripathi, L.~J. Guibas, and H.~Su,
  ``{PartNet}: A large-scale benchmark for fine-grained and hierarchical
  part-level {3D} object understanding,'' in {\em IEEE Conf. Comput. Vis.
  Pattern Recog.}, 2019.

\bibitem{choy2019fully}
C.~Choy, J.~Park, and V.~Koltun, ``Fully convolutional geometric features,'' in
  {\em IEEE Conf. Comput. Vis. Pattern Recog.}, 2019.

\bibitem{choy2020high}
C.~Choy, J.~Lee, R.~Ranftl, J.~Park, and V.~Koltun, ``High-dimensional
  convolutional networks for geometric pattern recognition,'' in {\em IEEE
  Conf. Comput. Vis. Pattern Recog.}, 2020.

\bibitem{gwak2020gsdn}
J.~Gwak, C.~B. Choy, and S.~Savarese, ``Generative sparse detection networks
  for 3d single-shot object detection,'' in {\em Eur. Conf. Comput. Vis.},
  2020.

\bibitem{PointContrast2020}
S.~Xie, J.~Gu, D.~Guo, C.~R. Qi, L.~Guibas, and O.~Litany, ``Pointcontrast:
  Unsupervised pre-training for 3d point cloud understanding,'' in {\em Eur.
  Conf. Comput. Vis.}, 2020.

\bibitem{hou2020exploring}
J.~Hou, B.~Graham, M.~Nie{\ss}ner, and S.~Xie, ``Exploring data-efficient 3d
  scene understanding with contrastive scene contexts,'' in {\em IEEE Conf.
  Comput. Vis. Pattern Recog.}, 2021.

\bibitem{Song2016cvpr}
S.~Song and J.~Xiao, ``Deep sliding shapes for amodal 3d object detection in
  rgb-d images,'' in {\em IEEE Conf. Comput. Vis. Pattern Recog.}, 2016.

\bibitem{qi2017frustum}
C.~R. Qi, W.~Liu, C.~Wu, H.~Su, and L.~J. Guibas, ``Frustum pointnets for 3d
  object detection from rgb-d data,'' in {\em IEEE Conf. Comput. Vis. Pattern
  Recog.}, 2018.

\bibitem{he_mocov1}
K.~He, H.~Fan, Y.~Wu, S.~Xie, and R.~Girshick, ``Momentum contrast for
  unsupervised visual representation learning,'' in {\em IEEE Conf. Comput.
  Vis. Pattern Recog.}, 2020.

\bibitem{chen2020mocov2}
X.~Chen, H.~Fan, R.~Girshick, and K.~He, ``Improved baselines with momentum
  contrastive learning,'' {\em arXiv preprint arXiv:2003.04297}, 2020.

\bibitem{chen2020simple}
T.~Chen, S.~Kornblith, M.~Norouzi, and G.~Hinton, ``A simple framework for
  contrastive learning of visual representations,'' {\em arXiv preprint
  arXiv:2002.05709}, 2020.

\bibitem{chen2020big}
T.~Chen, S.~Kornblith, K.~Swersky, M.~Norouzi, and G.~Hinton, ``Big
  self-supervised models are strong semi-supervised learners,'' {\em arXiv
  preprint arXiv:2006.10029}, 2020.

\bibitem{wang2020DenseCL}
X.~Wang, R.~Zhang, C.~Shen, T.~Kong, and L.~Li, ``Dense contrastive learning
  for self-supervised visual pre-training,'' in {\em IEEE Conf. Comput. Vis.
  Pattern Recog.}, 2021.

\bibitem{Wuunsupervise}
Z.~Wu, Y.~Xiong, S.~X. Yu, and D.~Lin, ``Unsupervised feature learning via
  non-parametric instance discrimination,'' in {\em IEEE Conf. Comput. Vis.
  Pattern Recog.}, 2018.

\bibitem{elich20193dbevis}
C.~Elich, F.~Engelmann, T.~Kontogianni, and B.~Leibe, ``{3D-BEVIS}:
  Bird's-eye-view instance segmentation,'' {\em arXiv preprint
  arXiv:1904.02199}, 2019.

\bibitem{Engelmann20ICRA}
F.~Engelmann, T.~Kontogianni, and B.~Leibe, ``Dilated point convolutions: On
  the receptive field size of point convolutions on 3d point clouds,'' in {\em
  Int. Conf. Robotics \& Automation}, 2020.

\bibitem{Narita2019iros}
G.~Narita, T.~Seno, T.~Ishikawa, and Y.~Kaji, ``Panopticfusion: Online
  volumetric semantic mapping at the level of stuff and things,'' in {\em Proc.
  IEEE Int. Conf. Intelligent Robots Syst.}, 2019.

\end{thebibliography}

	\begin{IEEEbiography}[{\includegraphics[width=1in,height=1.25in,clip,keepaspectratio]{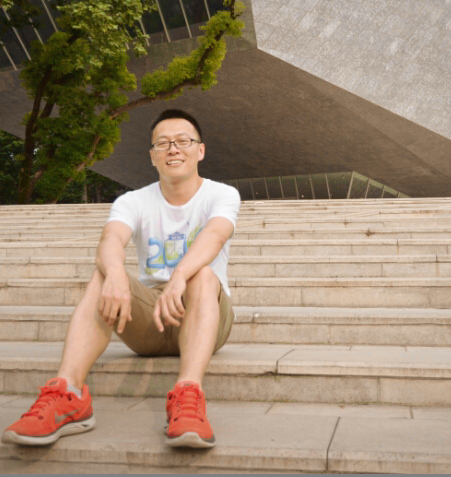}}]{Tong He}
		received the Ph.D. degree in computer science from the University of Adelaide, Australia in 2020. He is currently a researcher at Shanghai AI Laboratory. His research interests include computer vision and machine learning. 
	\end{IEEEbiography}
	
	\begin{IEEEbiographynophoto}%
 {Chunhua Shen}
  is a Chair  Professor at Zhejiang University. He was a Professor
  at The University of Adelaide, Australia.
	\end{IEEEbiographynophoto}
	
	
	\begin{IEEEbiography}[{\includegraphics[width=1in,height=1.25in,clip,keepaspectratio
 ]{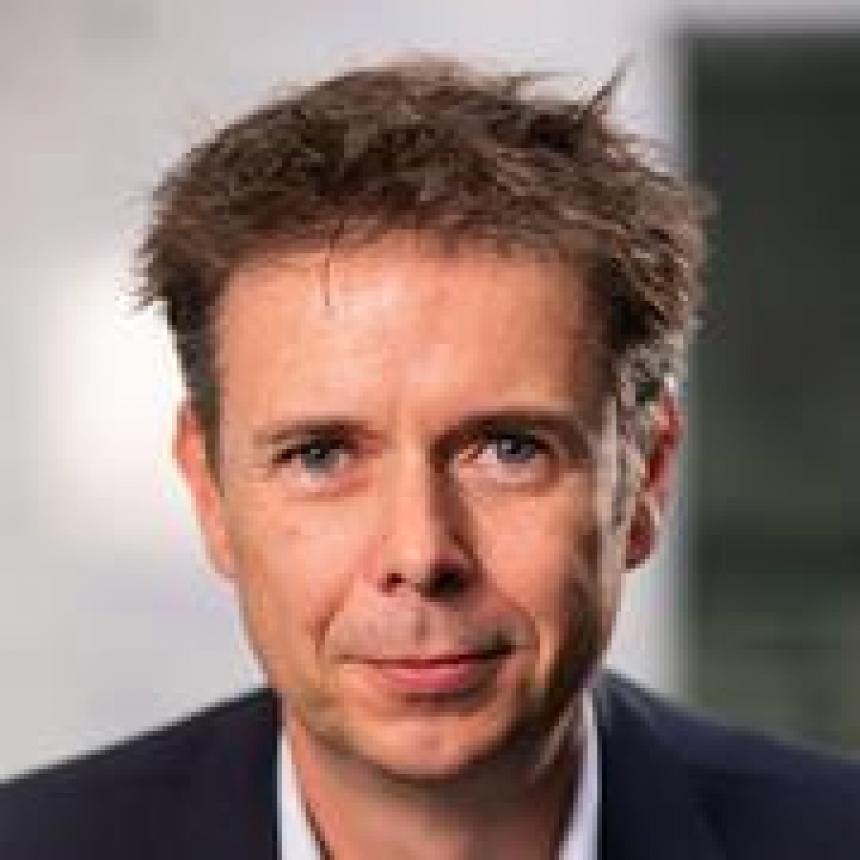}}]{Anton van den Hengel}
  is a Professor in The University of Adelaide.
	\end{IEEEbiography}

\end{document}